\documentclass[journal]{IEEEtran}

\usepackage{ifpdf}
\usepackage{color}
\usepackage{amssymb}
\usepackage[linesnumbered,ruled,vlined]{algorithm2e}
\usepackage{cite}
\usepackage{amsmath}
\usepackage{units}
\usepackage{array}
\usepackage{fixltx2e}
\usepackage{stfloats}
\usepackage{url}
\usepackage{hyperref}
\usepackage{subfigure}
\usepackage{footnote}
\usepackage{multirow}
\usepackage{booktabs}
\usepackage{tabularx}
\usepackage{makecell}

\usepackage[table,xcdraw]{xcolor}
\usepackage{threeparttable}
\usepackage{bbding}
\usepackage{orcidlink}

\begin{document}
\title{Towards Safe and Robust Autonomous Vehicle Platooning: A Self-Organizing Cooperative Decision-making Framework}

\author{Chengkai~Xu,
        Zihao~Deng,
        Jiaqi~Liu,
        Aijing~Kong,
        Yu~Tang,
        Chao Huang,~\IEEEmembership{Senior Member,~IEEE,} 
        and~Peng~Hang,~\IEEEmembership{Senior Member,~IEEE}%
\thanks{This work is supported in part by the State Key Laboratory of Intelligent Vehicle Safety Technology under Project No. IVSTSKL-202414, the National Natural Science Foundation of China (52302502), and the State Key Lab of Intelligent Transportation System under Project No. 2024-A002.}

\thanks{Chengkai Xu and Peng Hang are with the State Key Laboratory of Intelligent Vehicle Safety Technology and the College of Transportation, Tongji University, Shanghai 201804, China. (e-mail: xuchengkai@tongji.edu.cn, hangpeng@tongji.edu.cn)}

\thanks{
Zihao Deng is with the Institute of Automation, Chinese Academy of Sciences, Beijing 100190, China. (e-mail: dengzihao2025@ia.ac.cn)
}
\thanks{
Jiaqi Liu is with the Department of Computer Science, University of North Carolina at Chapel Hill, United States. (e-mail: jqliu@cs.unc.edu)
}
\thanks{Kong Aijing is with the School of Automotive Studies, Tongji University, Shanghai 201804, China. (e-mail: ajkong@tongji.edu.cn)}

\thanks{Yu Tang is with China Automotive Engineering Research Institute Co., Ltd, Chongqing 430058, China. (e-mail: tangyu@caeri.com.cn)}

\thanks{Chao Huang is with Department of Industrial and System Engineering, The Hong Kong Polytechnical University, Hong Kong 999077. (e-mail:  hchao.huang@polyu.edu.hk) }
\thanks{Corresponding author: Jiaqi Liu and Peng Hang}}

\markboth{IEEE Transactions on Vehicular Technology}%
{Shell \MakeLowercase{\textit{et al.}}: Bare Demo of IEEEtran.cls for IEEE Journals}

\maketitle

\begin{abstract}
In hybrid traffic environments where human-driven vehicles (HDVs) and autonomous vehicles (AVs) coexist, achieving safe and robust decision-making for AV platooning remains a complex challenge.
Existing platooning systems often struggle with dynamic formation management and adaptability, especially under complex and dynamic mixed-traffic conditions.
To enhance autonomous vehicle platooning within these hybrid environments, this paper presents TriCoD, a twin-world safety-enhanced Data-Model-Knowledge Triple-Driven Cooperative Decision-making Framework.
This framework integrates deep reinforcement learning (DRL) with model-driven approaches, enabling dynamic formation dissolution and reconfiguration through a safety-prioritized twin-world deduction mechanism. The DRL component augments traditional model-driven methods, enhancing both safety and operational efficiency, especially under emergency conditions. Additionally, an adaptive switching mechanism allows the system to seamlessly switch between data-driven and model-driven strategies based on real-time traffic demands, thus optimizing decision-making ability and adaptability. Simulation experiments and hardware-in-the-loop tests demonstrate that the proposed framework significantly improves safety, robustness, and flexibility. 
A detailed account of the validation results for the model can be found in
\href{https://perfectxu88.github.io/Towards/}{Our Website}.
\end{abstract}


\IEEEpeerreviewmaketitle

\section{Introduction}

\IEEEPARstart{T}{he} rapid advancements in autonomous driving and vehicle communication technologies have catalyzed significant interest in connected and autonomous vehicles (CAVs). Among these technological advances, autonomous vehicle platooning has emerged as a promising approach to enhance traffic efficiency, safety, and environmental sustainability \cite{zhu2021survey, fang2025towards}. By reducing the gaps between vehicles and aerodynamic drag, vehicle platooning holds substantial potential to decrease fuel consumption and emissions, address environmental challenges, and mitigate traffic congestion\cite{gautam2024overview}.

Current autonomous vehicle platooning systems primarily focus on two key aspects of vehicle control: Cooperative Adaptive Cruise Control (CACC) and Cooperative Lane Change Control (CLCC). CACC manages longitudinal dynamics to maintain inter-vehicle gaps and ensure string stability \cite{feng2019string}, while CLCC facilitates lateral maneuvers for lane changes within the platoon. Extensive research has been dedicated to improving CACC and CLCC. Model-based approaches, such as Model Predictive Control (MPC), have been employed to optimize vehicle dynamics \cite{nandhini2024comprehensive}. However, these approaches often struggle with computational complexity and adaptability to rapidly changing environments. Rule-based methods offer simplicity but lack flexibility \cite{wang2023driving, zhang2024efficient}, while Reinforcement Learning (RL) has recently emerged as a promising alternative capable of learning from real-time traffic data \cite{10919632, zhou2024enhancing}.

Despite these contributions, CACC and CLCC remain largely confined to fixed leader–follower chains and single-lane operation, which limits their ability to manage dynamic interactions and to reorganize a platoon in mixed traffic \cite{liu2024learning, braiteh2024platooning}. Recent work has begun to relax these assumptions. Hu \emph{et al.}~\cite{hu2024vehicles} treat platoon as intelligent swarm, coupling longitudinal and lateral cooperation through multi-agent planning, whereas Zhou \emph{et al.}~\cite{zhou2024enhancing} embed safe-RL shields to mitigate the stochastic behavior of human-driven vehicles (HDVs). Although both studies demonstrate that vehicle–vehicle cooperation can indeed be made more adaptive and safer, their formations remain logically intact and neither framework supports explicit dissolution, re-merging, or topology switching when the traffic context changes.

To overcome these challenges, inspiration is drawn from formation decision-making techniques in other domains, specifically Unmanned Aerial Vehicles (UAVs). In the broader context of multi-agent systems, the concept of formation reconfiguration has been extensively explored\cite{li2025deform}, which leverage dynamic formation reconfiguration to adapt to changing conditions, avoid obstacles, and fulfill mission requirements in controlled three-dimensional environments\cite{tian2024formation}. Although UAVs operate in different environments compared to ground vehicles, certain principles, such as decentralized control and dynamic adaptability, can be applied to autonomous vehicle platoons to improve their formation decision capabilities.

Unlike UAVs, which operate in three dimensions with relatively low obstacle densities, autonomous vehicle platoons must navigate constrained road environments with frequent interactions involving HDVs and complex infrastructure. The unpredictable behavior of HDVs, combined with road constraints, presents a significant challenge for dynamic formation reconfiguration in vehicle platoons\cite{xu2025survey, zhou2023cooperative}. Addressing these challenges requires a novel approach that integrates adaptability with the specific needs of ground-based traffic systems.

In response, this paper proposes TriCoD, a safety-enhanced Data-Model-Knowledge \textbf{Tri}ple-Driven Adaptive \textbf{Co}operative \textbf{D}ecision-making Framework for autonomous vehicle platooning, integrating data-driven methods with model-based control to manage platoons in complex environments. The platooning problem is modeled as a cooperative decision-making challenge, with a twin-world safety enhancement model running in parallel with real scenarios to improve safety and adaptability during learning. An adaptive switching mechanism optimizes the transitions between traffic conditions-based decision strategies.

\begin{figure}
    \centering
    \includegraphics[width =0.48\textwidth]{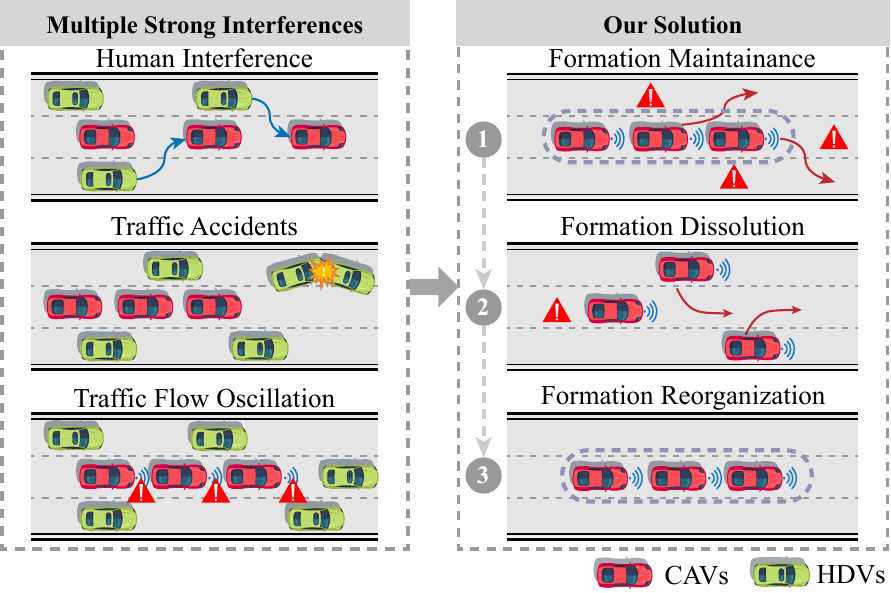}
    \caption{Illustration of challenging traffic conditions and our framework's adaptive responses. The diagram demonstrates the dynamic stages of (1) maintaining formation with active avoidance, (2) realigning to mitigate the impact of disruptions, and (3) restoring optimal formation after the disturbances, with CAVs (red) and HDVs (green) represented with corresponding behaviors under each scenario.}
    \label{fig:traffic scenario}
\end{figure}

The hybrid approach is designed to enable platoon reconfiguration in complex traffic conditions. The system dynamically adjusts both longitudinal and lateral behaviors to enhance safety, efficiency, and flexibility in real time.
The approach is evaluated against structured disturbances in mixed-traffic high-risk scenarios, including human interference, incidents, and fluctuating flows, as shown in Figure \ref{fig:traffic scenario}.
In these scenarios, the process of platoon reorganization plays a critical role in enhancing the system's safety and adaptability. Our framework demonstrates the ability to dynamically dissolve and reorganize the platoon formation in response to disturbances, ensuring both safety and operational efficiency. Extensive software-in-the-loop and hardware-in-the-loop experiments fully demonstrate that our framework exhibits higher security and robustness in different scenarios while taking efficiency into consideration.

The contributions of this paper are summarized as follows:

\begin{itemize}
    \item A hybrid framework for platooning decision has been developed, which integrates data-driven methods with model-based control and domain-specific knowledge, enabling platoon members to achieve adaptive decision-making in complex traffic environments.
    \item A twin-world safety enhancement model integrated with RL is proposed, which significantly enhances sampling efficiency by providing a parallel virtual environment that allows safe exploration of diverse strategies, thereby improving the overall safety and adaptability of the system.
    \item The proposed framework uniquely enables the dissolution and reconfiguration of platoon formations in response to changing traffic conditions. It integrates longitudinal and lateral control, allowing the system to dynamically adjust vehicle behavior and maintain formation integrity in complex environments.
\end{itemize}

To better present our work, the rest of this paper is organized as follows. \textit{Section \ref{Problem statement}} formulates the problem in our research.
\textit{Section \ref{Methodology}} introduces in detail the proposed TriCoD framework.
\textit{Section \ref{Simulation and Performance Evaluation}} presents the performance evaluation.
Finally, the paper is concluded in \textit{Section \ref{Conclusion}}.

\section{Related Works}
\subsection{Coordinated Lane Change Decision-Making}

Coordinated lane change allows a platoon to merge, overtake, or circumvent obstacles while preserving safety and throughput. 
The controller must determine both the instant and path of the manoeuvre under tight kinematic, traffic-flow, and regulatory constraints, which is an intrinsically difficult optimization problem.  
The literature addresses this task with \emph{model-based}, \emph{rule-based}, and \emph{data-driven} paradigms.

\paragraph*{Model-based control}
Optimal-control formulations such as Model Predictive Control and Linear-Quadratic Regulation solve a constrained optimization at each step to ensure string stability and collision avoidance\cite{wei2024ensuring, yao2024planning}.
These methods offer the advantage of providing clear performance guarantees and are computationally efficient, yet their efficacy hinges on accurate dynamical models and reliable previews of surrounding traffic\cite{zhang2023platoon}. In mixed traffic containing HDVs, model mismatch and sensing uncertainty can erode safety margins.

\paragraph*{Rule-based heuristics}
Rule-based strategies embed expert knowledge in a finite set of heuristics that trigger lane changes when pre‑specified conditions are met\cite{wang2023driving, ma2023collision}. These rules typically consider factors like relative velocity, gap acceptance, and the presence of obstacles. While such systems are straightforward to implement and interpret, they lack the adaptability required to respond to complex and unforeseen traffic situations. The rigid nature of rule-based approaches can lead to suboptimal performance in scenarios that deviate from the predefined conditions.

\paragraph*{Data-driven learning}
Recent advances have seen the integration of data-driven techniques, particularly RL, to enhance lane change decision-making\cite{zhou2024enhancing, xu2025tell}. RL allows agents to learn strategies through interaction with the environment, which has demonstrated an ability to capture nonlinear vehicle interactions and to negotiate complex, uncertain environments that foil classical controllers, yet faces slow convergence, sensitivity to reward shaping, and the absence of formal safety guarantees during exploration and deployment\cite{jiang2023learning, peng2024efficient}.

\paragraph*{Hybrid directions and open gaps}
However, each paradigm exhibits inherent deficiencies: model‑based schemes struggle with uncertainty, rule‑based logic lacks adaptability, and purely data‑driven solutions face stability and safety barriers. Recent work has begun to explore hybrid frameworks that fuse analytical models with learning or heuristic layers\cite{gao2025virtual}. However, research in this direction remains nascent, where key questions include how to orchestrate the blending of components, provide formal safety guarantees, and maintain real‑time feasibility. These gaps motivate the mixed architecture proposed in this study.

\subsection{Formation Reorganization Decision-Making}
Most platoon controllers maintain a rigid leader–follower chain and rely on longitudinal regulation such as CACC and isolated lateral manoeuvres via CLCC. Such static topologies cannot dissolve, split, or reform in real time, limiting resilience to incidents and traffic oscillations\cite{braiteh2024platooning, 10919632}.

\paragraph*{Insights from robotics and UAV swarms}
Multi-robot teams and UAV swarms routinely perform formation reconfiguration to bypass obstacles, adapt to agent loss, or satisfy mission geometry\cite{gao2023hybrid, park2023formation}.  
Decentralized consensus, virtual-structure, and behavior-based methods deliver flexibility and robustness in three-dimensional free space.  
However, these algorithms assume near-perfect observability and lack the lane level and legal constraints of ground traffic.

\paragraph*{Emerging vehicular studies}
Recent work has begun to translate reorganization concepts to road traffic, exploring dynamic leader selection and splitting or merging of multilane platoon under mixed autonomy. 
Although simulation results show promise, most studies presume homogeneous CAVs and ideal vehicle-to-vehicle (V2V) communication, leaving their robustness in dense traffic unverified\cite{hu2024vehicles}.

\paragraph*{Domain-specific challenges}
Vehicles operate on segmented roadways with severe kinematic limits and must comply with traffic regulations.  
Formation reorganization on roads requires longitudinal–lateral coordination, safety guarantees during collective manoeuvres and resilience to non-cooperative HDVs, which are not fully met by existing frameworks\cite{zhou2023cooperative}.

To date, there is a lack of comprehensive architecture that delivers \emph{verifiably safe}, \emph{computationally tractable}, and \emph{adaptive} in mixed traffic.
Therefore, this paper is dedicated to applying the formation reorganization technology to autonomous driving formations and building a unified decision-making layer that combines lane splitting, merging, and lane changing planning in mixed traffic flows.

\section{Problem statement}
\label{Problem statement}
\subsection{Scenario Description}

Highway environments are the most common application scenario for autonomous vehicle platooning. This study focuses on self-organizing decision-making for multi-vehicle platooning on a three-lane highway, addressing significant disturbances such as human interference, traffic accidents, and traffic flow oscillations, as illustrated in Figure \ref{fig:traffic scenario}. In these scenarios, traditional formation algorithms are no longer capable of effectively addressing the challenges posed by unpredictable disturbances, leading to reduced stability and compromised safety. Therefore, a new, adaptive algorithm is necessary to navigate these disturbances and ensure robust, stable, high-speed operation.

\subsection{Problem Formulation}
\begin{figure}[t]
        \centering
        \includegraphics[width = 0.4\textwidth]{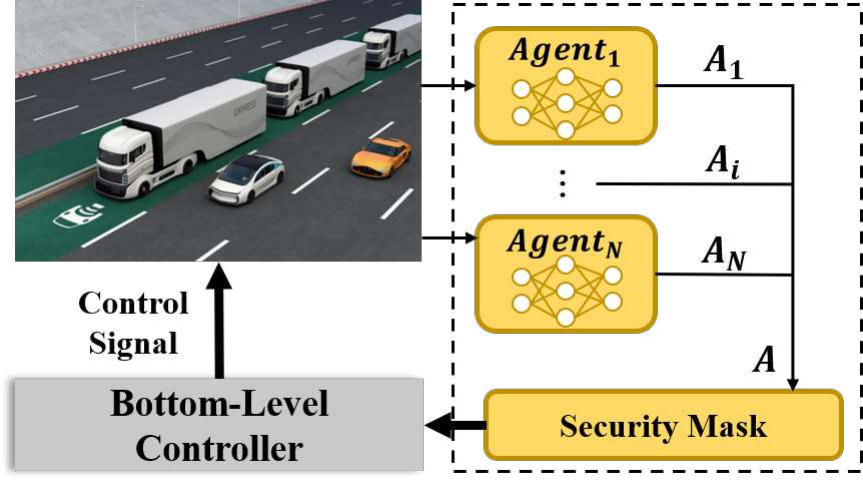}
        \caption{Illustration of the system and the simulation setup. All high-level actions are verified through a security mask before being processed by the bottom-level controller, which will be finally converted into control signals.}
        \label{schematics of the system}
\end{figure}

The problem is modeled as a Partially Observable Markov Decision Process (POMDP)\cite{spaan2012partially}, which provides a robust framework for handling the partial observability and stochastic nature of real-world traffic scenarios. By employing DRL within the POMDP framework, both individual and collective goals can be balanced, enabling adaptive responses in dynamic environments. The system schematics are shown in Figure \ref{schematics of the system}.

The vehicle platoon, while making decisions as a whole, employs a decentralized execution strategy. The problem is formulated as a POMDP, which can be described by the tuple \((\mathcal{S}, \mathcal{A}, \mathcal{P}, \mathcal{R}, \mathcal{O})\), where \(\mathcal{S}\) is the set of system states; \(\mathcal{A}\) is the action space; \(\mathcal{P}\) is the state transition model; \(\mathcal{R}\) is the reward function; and \(\mathcal{O}\) is the observation space. 

In this POMDP, the agent makes decisions based on the policy \( \pi \) and the current observation space. The optimization problem is then formulated as finding the optimal policy \( \pi^* \) that maximizes the expected sum of discounted future rewards\cite{liu2023towards}: 

\begin{equation}
    Q^{\pi}(s, a) \stackrel{\text{def}}{=} \mathbb{E}_{\pi} \left[ \sum_{t=0}^{\infty} \gamma^t R(s_t, a_t) \mid s_0 = s, a_0 = a \right]
\end{equation}
where \( \gamma \in [0,1] \) is the discount factor that balances the importance of immediate versus future rewards. The optimization process involves updating the policy \( \pi_\theta \) to approximate the optimal policy \( \pi^* \), which governs how the platoon should act to achieve long-term performance goals.

\subsubsection{\textbf{Observation Space \( \mathcal{O} \)}}

In real-world highway environments, autonomous vehicles cannot directly observe the full state of the system due to sensor limitations and occlusions caused by other vehicles. Formally, the observation \( o_t \) at time \( t \) is a function of the true state \( s_t \), but it is incomplete:

\begin{equation}
    o_t = f(s_t), \quad o_t \subset s_t
\end{equation}

In this study, the observation space is structured as a matrix with dimensions \(|N_i| \times |F|\), where \(|N_i|\) is the number of vehicles detected by the platoon, and \(|F|\) represents the feature vector describing each detected vehicle. The feature vector includes relative positions and velocities of other vehicles, as well as an indicator variable \( \delta_i^{platoon} \) that specifies whether the vehicle is part of the platoon:

\begin{equation}
    F_i = [x_i, y_i, v_i^x, v_i^y, \delta_i^{platoon}]
\end{equation}
where \( x_i \) and \( y_i \) represent the relative longitudinal and lateral positions of vehicle \( i \), and \( v_i^x \) and \( v_i^y \) denote its relative longitudinal and lateral velocities. The binary variable \( \delta_i^{platoon} \) indicates whether vehicle \( i \) is part of the platoon:

\begin{equation}
    \delta_i^{platoon} =
    \begin{cases}
        1 & \text{if vehicle } i \text{ belongs to the platoon}, \\
        0 & \text{otherwise}.
    \end{cases}
\end{equation}

Additionally, the observation is restricted to vehicles within a sensor detection range, \( D_{vision} \), ahead or behind the observing vehicle.

\subsubsection{\textbf{Action Space \( \mathcal{A} \)}}

The action space \( \mathcal{A} \) defines the set of possible actions that the platoon can take at any given time step. In this study, each autonomous vehicle in the platoon selects from a discrete set of high-level actions, including \textit{left turn, right turn, accelerate, decelerate, and cruise}. These actions allow the vehicles to maneuver within their lanes and adjust their speeds to maintain a safe and efficient formation.

For a platoon consisting of \( N \) autonomous vehicles, the joint action space \( \mathcal{A} \) is the Cartesian product of the individual action spaces of the vehicles:

\begin{equation}
    \mathcal{A} = \mathcal{A}_1 \times \mathcal{A}_2 \times \cdots \times \mathcal{A}_N
\end{equation}

Each vehicle \( i \) in the platoon selects an action \( a_i \in \mathcal{A}_i \), and the combined action \( a_{platoon} \) is executed as:

\begin{equation}
    a_{platoon} = (a_1, a_2, \ldots, a_N)
\end{equation}

In this study, a quinary encoding scheme is employed to encode the joint action space, resulting in a 125-dimensional discrete action space through the Cartesian product of the individual actions. After selecting high-level actions, the platoon's low-level controllers manage steering and throttle to execute the specific vehicle movements in the simulation environment, as illustrated in Figure \ref{schematics of the system}.

\subsubsection{\textbf{Transition Probabilities \( \mathcal{P} \)}}
HDVs introduce randomness into the whole system, making future states inherently uncertain. This uncertainty is modeled using a state transition function \( \mathcal{P}(s_{t+1} | s_t, a_t) \), which represents the stochastic nature of the environment, capturing the likelihood of transitioning from the current state \( s_t \) to the next state \( s_{t+1} \) given the current action \( a_t \):

\begin{equation}
    \mathcal{P}(s_{t+1} | s_t, a_t) = \Pr(s_{t+1} | s_t, a_t)
\end{equation}

The transition probabilities account for the stochasticity brought by the HDVs, ensuring that the decision-making process anticipates and responds to these uncertainties.

\subsection{Vehicle Model}

The vehicle kinematics model is commonly used for decision-making and planning in autonomous driving. Therefore, this paper adopts a simplified nonlinear bicycle model for the DRL-based self-organizing cooperative decision-making, described by
\begin{equation}
\begin{aligned}
\dot{x} &= v \cdot \cos(\psi + \beta) \\
\dot{y} &= v \cdot \sin(\psi + \beta) \\
\dot{\psi} &= \frac{v}{l} \cdot \sin \beta \\
\beta &= \tan^{-1}\left(1/2 \cdot \tan\delta\right)
\end{aligned}
\end{equation}
where $(x, y)$ is the global position, $v$ is vehicle velocity, $\beta$ is the slip angle at the vehicle’s center of gravity due to steering, $\psi$ denotes the yaw angle, $l$ denotes the wheelbase, and $\delta$ denotes the front steering angle.

This system enables the vehicle to adapt its position and orientation to follow a desired path with both precision and stability, which underpins both the DRL and the linear design of the LQR in \textit{Section \ref{Methodology}}.

\section{Methodology}
\label{Methodology}

This section introduces a comprehensive framework for autonomous vehicle platoon. The DRL-based core decision-making algorithm for formation reconfiguration is first presented to enable adaptive and secure platoon adjustments, which is followed by a formation maintenance algorithm and an adaptive switching mechanism, ensuring stable and efficient decision-making across varying traffic conditions.

\subsection{Framework Overview}

\begin{figure*}
    \centering
    \includegraphics[width = \textwidth]{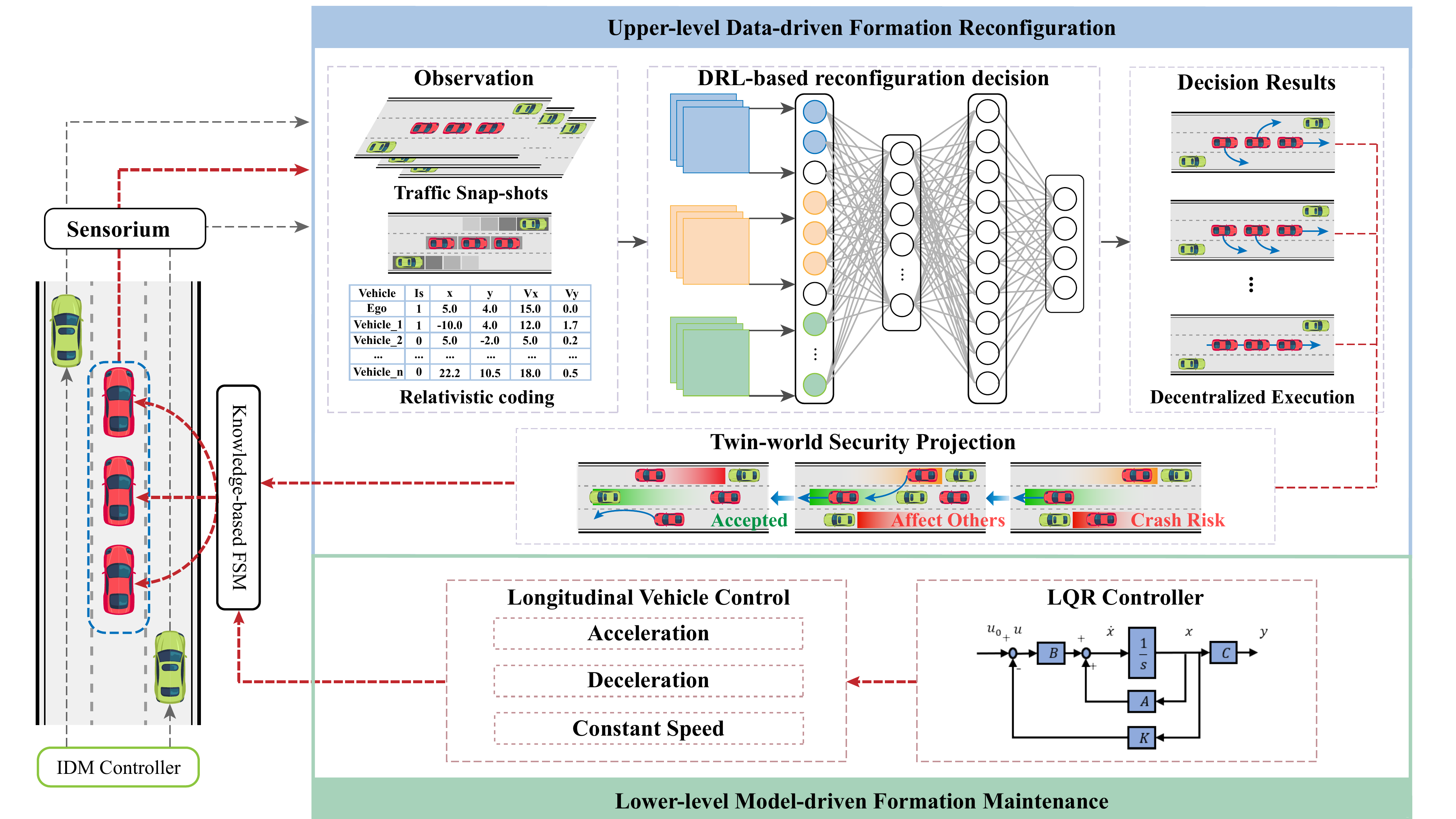}
    \caption{Overview of the TriCoD framework for self-organizing autonomous vehicle platooning, which combines a data-driven upper layer and a model-driven lower layer to manage platoon adaptively. The upper layer uses DRL with a twin-world safety verification model for safe, real-time decision-making, while the lower layer employs a LQR for precise control in predictable conditions, ensuring stable and efficient vehicle spacing. Together, these layers provide robust, adaptive platooning in complex traffic environments.}
    \label{overview}
\end{figure*}

To address emergency and high-risk scenarios that traditional vehicle platooning technologies struggle to manage effectively, this study introduces TriCoD framework. As depicted in Figure \ref{overview}, the framework seamlessly integrates two tightly coupled primary layers that collaboratively handle different aspects of platoon decision making:

\begin{itemize}
    \item \textbf{Upper Layer:} The DRL decision model follows the Centralized Training and Decentralized Execution (CTDE) paradigm, with a twin-world safety verification model that simulates decision outcomes in a parallel virtual environment. This verification ensures safe, effective decision-making before deployment and enhances the DRL model’s sampling efficiency. The upper layer works alongside the lower layer to adapt to complex scenarios in real-time.
    \item \textbf{Lower Layer:} The Linear-Quadratic Regulator (LQR) provides precise control in predictable scenarios, minimizing headway for efficiency. This layer also collaborates with the DRL model by executing low-level control commands that align with the strategic decisions made by the upper layer.
\end{itemize}

A knowledge-based finite state machine dynamically integrates these layers, switching decision strategies based on real-time traffic conditions. In stable conditions, the LQR controller optimizes efficiency with reduced headway. In complex scenarios, the DRL model manages platoon splitting and reformation, coordinating with the LQR for adaptive responses. This hybrid approach combines data-driven learning with model-based control, ensuring cohesive, adaptive responses to both routine and emergency situations. 

\subsection{Reward for Platooning}
The reward function is central to shaping the behavior of autonomous vehicle platoons. Inspired by the social behavior in AVs\cite{toghi2022social}, a reward system is designed to account for both individual vehicle performance and the collective dynamics of the platoon as a whole. This approach ensures that each vehicle operates safely and efficiently while maintaining coordination and cohesion within the platoon.

The total reward \( R_{global}(s,a) \) is formulated as the sum of individual rewards and system-wide collaborative rewards:

\begin{equation}
    R_{global}(s,a) =  \frac{1}{n}\sum_{i=1}^n R_{ind}^{i} + R_{sys}
\end{equation}

Both the individual vehicle perspective and the platoon-wide perspective are considered in the reward design. From the individual vehicle’s perspective, the reward function emphasizes safety and efficiency, focusing on collision avoidance, lane keeping, speed control, and smooth acceleration:

\begin{equation}
    R_{ind}(s,a) = \sum_{t \in \{C,L,F,A\}} w_{t} \times r_{t}
\end{equation}
where \( r_C \) penalizes collisions, \( r_L \) encourages lane discipline, \( r_F \) rewards maintaining high speeds, and \( r_A \) controls acceleration behavior. The weights \( w_C \), \( w_L \), \( w_F \), and \( w_A \) are tuned to reflect the importance of each factor, with a strong emphasis on collision avoidance through a higher value of \( w_C \). 

Each component is defined as follows:

\begin{equation}
\left\{
\begin{aligned}
    r_C &= -\mathbb{I}_{\{collision\}} \\
    r_L &= - cos(\pi (y - y_{center})/4) \\
    r_F &= clip\left( (v - v_{\min})/(v_{\max} - v_{\min}),\ 0,\ 1 \right)\\
    r_A &= 1-|{a}/{a_{max}}| 
\end{aligned}
\right.
\label{eq:individual_rewards}
\end{equation}
where \( v \) denotes longitudinal speed, \( a \) acceleration, and \( y \) lateral position. The operator \(\mathbb{I}_{\{\cdot\}}\) denotes the indicator function, and $clip$ restricts the value within \([0,1]\) to stabilize learning.
From the platoon-wide perspective, the reward function promotes the cooperation and cohesion of the entire platoon. The system-wide reward \( R_{sys}(s,a) \) encourages vehicles to remain coordinated and maintain proper formation:

\begin{equation}
    R_{sys}(s,a) = \sum_{k \in \{M,H,S\}} w_{k} \times r_{k}
\end{equation}

In this expression, \( r_M \) rewards staying in the same lane, \( r_H \) ensures proper headway between vehicles, and \( r_S \) aligns speed coordination within the platoon. The corresponding weights \( w_M \), \( w_H \), and \( w_S \) are designed to ensure that each vehicle's actions contribute to the collective success of the formation.

\begin{equation}
\left\{
\begin{aligned}
    r_M &= 1-n_{lane}/n_{max} \\
    r_H &= -\frac{1}{N-1} \sum_{i=1}^{N-1} \left| d_{i,i+1} - d^* \right| \\
    r_S &= -\frac{1}{N} \sum_{i=1}^{N} \left| v_i - \bar{v} \right|
\end{aligned}
\right.
\label{eq:system_rewards}
\end{equation}
where $n_{lane}$ and $n_{max}$ denotes the number of lanes occupied and the total number of lanes respectively. \( d_{i,i+1} \) is the inter-vehicle gap between vehicle \( i \) and its immediate follower, with \( d^* \) representing the target headway. \( v_i \) is the velocity of vehicle \( i \), and \( \bar{v} = \frac{1}{N} \sum_{i=1}^{N} v_i \) is the average platoon speed.

\begin{figure}
    \centering
    \includegraphics[width=0.8\linewidth]{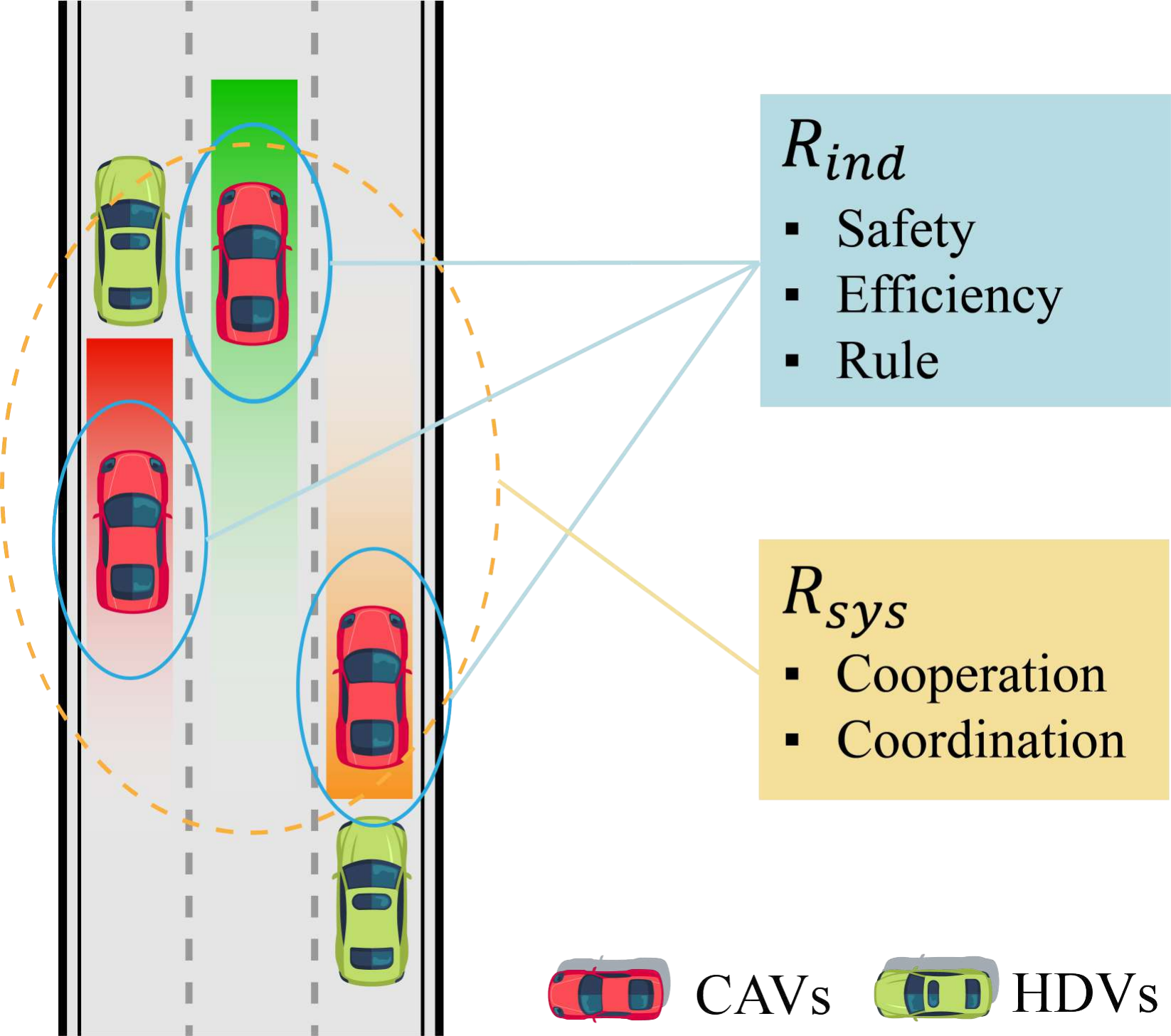}
    \caption{Illustration of the designed headway maintaining reward, which considers both the single-vehicle rewards and the multi-vehicle rewards.}
    \label{reward stuc}
\end{figure}

This reward design effectively balances the individual objectives of safety and efficiency with the collective need for coordination, ensuring that the platoon operates as a cohesive unit. The combined rewards guide both individual vehicles and the platoon as a whole toward achieving optimal performance in dynamic traffic conditions. Figure \ref{reward stuc} illustrates the structure of the reward function, highlighting how it integrates both individual and collective factors.

\subsection{Deep Reinforcement Learning Algorithm}
In DRL, an agent interacts with an environment defined by a Markov Decision Process (MDP), aiming to learn a policy that maximizes cumulative long-term rewards. Value-based DRL methods estimate the state–action value function \( Q(s,a) \) and obtain the optimal action by \( a^* = \arg\max_a Q(s,a) \). 
However, in multi-vehicle platooning, the joint action space grows exponentially with the number of agents, and Q-learning suffers from poor sample efficiency, instability due to non-stationarity from concurrent updates, and brittle off-policy corrections that destabilize closely coupled dynamics.

To address these issues, we adopt a policy gradient approach based on an actor–critic architecture, which separates the representation of the policy (actor) from that of the value function (critic). The actor directly maps observed states to a distribution over actions, while the critic provides a learned estimate of the state value \( V^\pi(s) \) and the advantage function \( A^\pi(s,a) \), allowing a reduction of the variance of the gradient estimation. The target of agent is to optimize the long-term cumulative reward under policy \(\pi\):

\begin{equation}
    J(\pi) = \underset{\pi}{argmax} \quad \mathbb{E}_{\pi} \left[ \sum_{t=0}^{T} \gamma^t r(s_t,a_t) \right]
\end{equation}
where \( \gamma \in [0,1) \) is the discount factor and \(T\) is the time horizon. Using the policy gradient theorem, the gradient of the objective can be expressed as:
\begin{equation}
    \nabla_\theta J(\pi_\theta) = \mathbb{E}_{\pi_\theta} \left[ \nabla_\theta \log \pi_\theta(a_t|s_t) A_t^{\pi} \right]
\end{equation}
where \( A_t^{\pi} = Q^\pi(s_t,a_t) - V^\pi(s_t) \) is the advantage function. The advantage estimates are constructed from the Bellman equation, where the state value function \(V^\pi(s_t)\) and Q-value function \(Q^\pi(s_t, a_t)\) can be computed as:
\begin{equation}
    V^\pi(s_t) = \mathbb{E}_{a_t \sim \pi(\cdot|s_t)} \left[Q^\pi(s_t, a_t)\right]
\end{equation}
\begin{equation}
    Q^\pi(s_t, a_t) = r_t + \gamma \mathbb{E}_{s_{t+1} \sim T(\cdot|s_t,a_t)} \left[ V^\pi(s_{t+1}) \right]
\end{equation}

To ensure the stability of policy gradients, Trust Region Policy Optimization (TRPO) is introduced to ensure monotonic performance improvement and bounded divergence between successive policies \cite{schulman2015trust}:
\begin{equation}
\begin{aligned}
    \max_{\theta} \ \mathbb{E}_{\pi_{\theta_{old}}} \left[ \frac{\pi_\theta(a|s)}{\pi_{\theta_{old}}(a|s)} A^{\pi_{\theta_{old}}}(s,a) \right] \\
    \mathbb{E}_{s \sim \pi_{\theta_{old}}} \left[ KL \left( \pi_{\theta_{old}}(\cdot|s) \,\|\, \pi_{\theta}(\cdot|s) \right) \right] \leq \delta
\end{aligned}
\end{equation}
where \( \delta \) is a small positive constant controlling the size of the update step, and KL denotes the Kullback-Leibler divergence, which guarantees stable learning for multi-agent platooning.

\begin{figure}
    \centering
    \includegraphics[width = 0.4\textwidth]{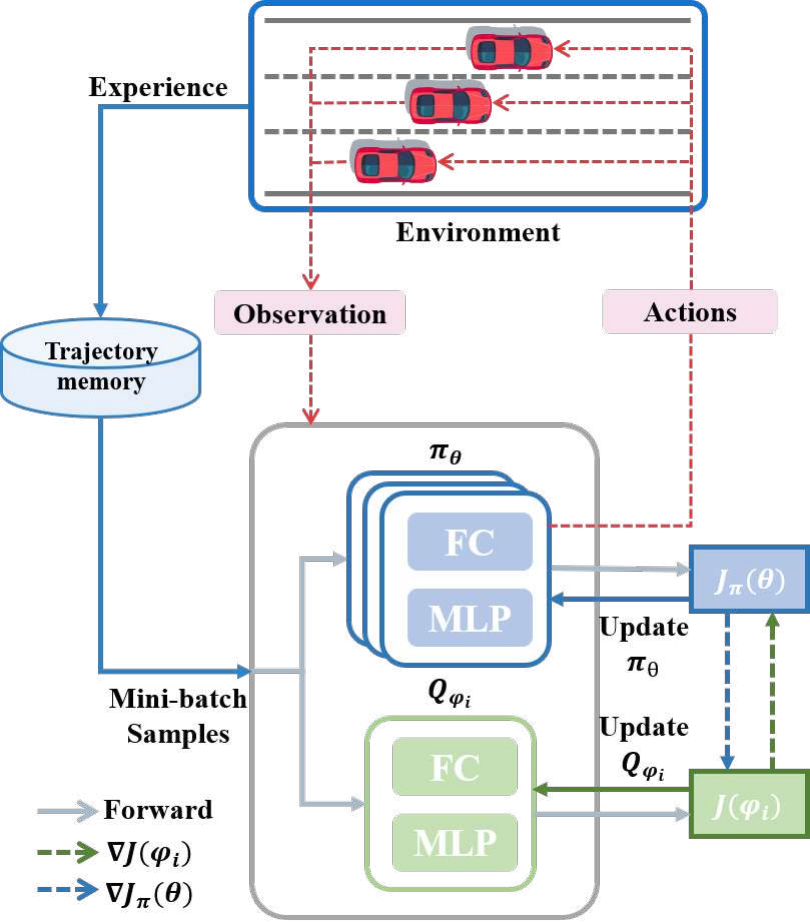}
    \caption{Illustration of the TRPO-based formation configuration algorithm schematic model.}
    \label{TRPO-based formation}
\end{figure}

In this study, a CTDE strategy shown in Figure~\ref{TRPO-based formation} is adopted. During training, sensor data from all platoon members are aggregated to form a global observation space. Then a shared policy \( \pi_\theta \) is learned using the TRPO algorithm. During execution, each agent applies the learned policy using only its local observations, ensuring decentralized operation and scalability.

\begin{figure*}
    \centering
    \includegraphics[width = 0.9\textwidth]{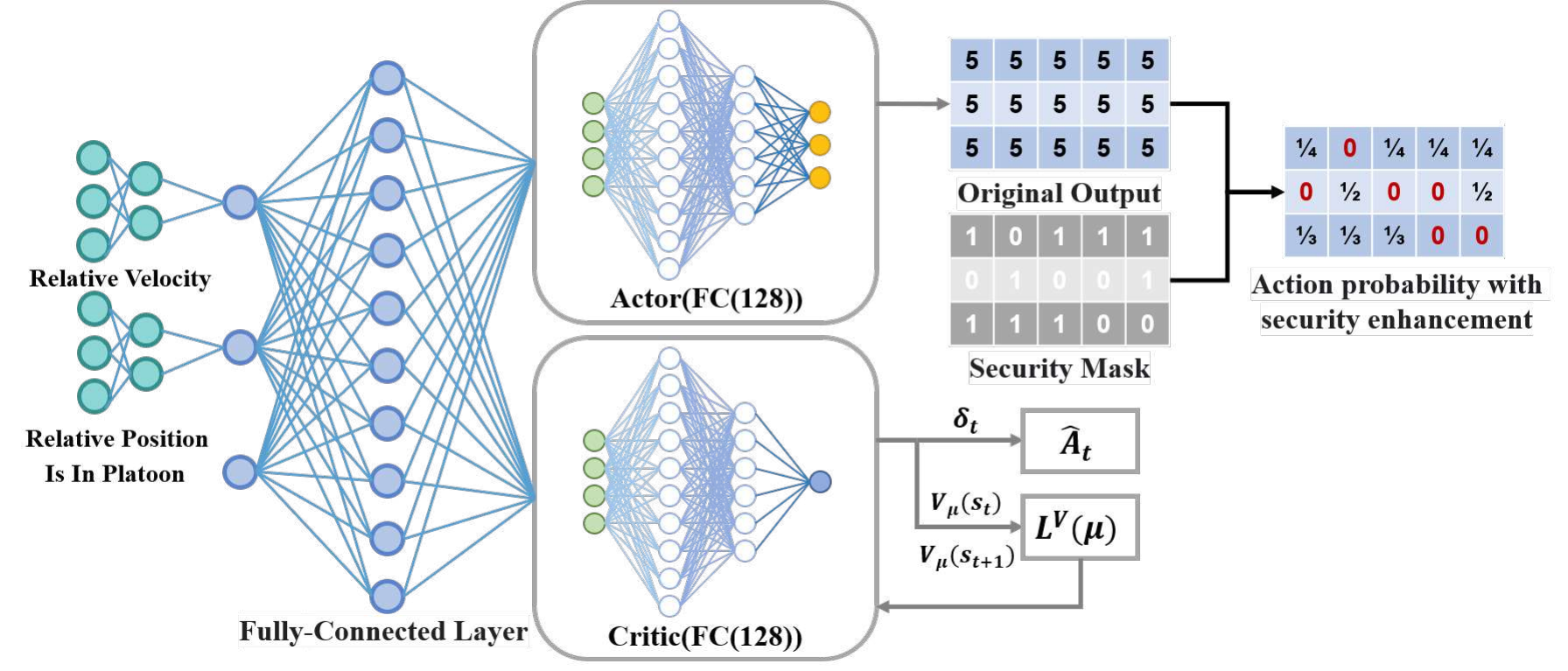}
    \caption{Illustration of the Actor-Critic Network Structure with Security-Enhanced Action Probability, which processes inputs through full connected layers in the Actor-Critic networks. The Actor generates action probabilities, which are refined by a security mask to ensure safe decisions, while the Critic evaluates actions to maximize cumulative rewards.} 
    \label{the network structure}
\end{figure*}

Since \(V^\pi(s_t)\) cannot be calculated directly, the improved deep neural networks (DNN) shown in Figure \ref{the network structure} is adopted to approximate the actor and critic, which processes input states through fully connected layers. The actor outputs action probabilities that are refined with a safety mask to eliminate unsafe actions. The critic evaluates state values to support policy improvement.

In this architecture, the policy loss and value function error loss are combined into a single loss function. Under the condition of shared parameters among agents, the total loss function can be expressed as a linear combination of the policy loss, state update loss, and entropy regularization term:

\begin{equation}
    J\left(\theta\right) = J^{\pi_{\theta}} - \beta_1 J^{V_{\phi}} + \beta_2 H\left(\pi_{\theta}\left(s_t\right)\right)
\end{equation} 
where \(\beta_1\) and \(\beta_2\) are the weight values for the value loss function and the entropy regularization term, respectively. The policy loss \( J^{\pi_{\theta}} \) and the state update loss function \( J^{V_{\phi}} \) are defined as:

\begin{equation}
    J^{\pi_{\theta}} = \mathbb{E}_{\pi_{\theta}} \left[ \log \pi_{\theta} \left( a_{t} | s_{t} \right) A_{t}^{\pi_{\theta}} \right]
\end{equation}

\begin{equation}
    J^{V_{\phi}} = \min_{\phi} \mathbb{E}_{\mathcal{D}} \left[ \left( r_{t} + \gamma V_{\phi} \left( s_{t+1} \right) - V_{\phi} \left( s_{t} \right) \right) \right]^2
\end{equation}
where \( A_{t}^{\pi_{\theta}} \) is the advantage function, defined as:

\begin{equation}
\label{advantage function}
    A_{t}^{\pi_{\theta}} = r_{t} + \gamma V^{\pi_{\phi}} (s_{t+1}) - V^{\pi_{\phi}} (s_{t})
\end{equation}
where \( V^{\pi_{\phi}} (s_{t}) \) is the state value function. Additionally, the entropy regularization term , which encourages the agent to explore new directions and avoid potential local optima, can be expressed as:

\begin{equation}
    H\left(\pi_{\theta}\left(s_t\right)\right) = \mathbb{E}_{\pi_{\theta}} \left[ -\log \left(\pi_{\theta}\left(s_t\right)\right) \right]
\end{equation}

Mini-batch stochastic gradient descent is used to update the network parameters. The use of the trust-region constraint ensures that updates remain conservative, thereby preserving policy stability. Together with the twin-world safety verification (detailed in \textit{Section \ref{Twin_World_Security_Projection}}), this setup allows the agent to safely explore reconfiguration strategies under diverse traffic conditions.

\subsection{Twin World Security Projection}
\label{Twin_World_Security_Projection}
While DRL enables robust decision-making in mixed traffic with strong interferences, its "black-box" nature can lead to unexpected outcomes in standard scenarios and may not fully comply with traffic regulations. Although more reward functions could enforce these constraints, they may incur significant training costs and risk model divergence. These issues are addressed in this study through the implementation of a comprehensive and extensible safety mask.

To ensure decision safety and sample efficiency, a safety-priority-based twin-world model is proposed. This model evaluates the safety of DRL decisions by first detecting potential collisions based on safety priorities, thereby reducing computational complexity. It then recursively computes \(T_n\) steps in a twin world constructed from the current scenario, optimizing actions based on the platoon's safety margin.

Given the focus of this study on decision-making in mixed traffic, predicting the driving behavior of HDVs is essential.  The prevalent Intelligent Driver Model (IDM) \cite{treiber2000congested} is utilized for simulating car-following behaviors, while the MOBIL model \cite{kesting2007general} is adopted for lane-changing decisions. The combination of these models minimizes collision risk among HDVs while closely mirroring human driving patterns.

The safety mask models each vehicle as a circle, with the center at the vehicle's center and a radius equal to half its length. Future \( T_n \) steps are simulated using motion models of HDVs and CAVs to assess collision risks. While this simulation is straightforward for individual vehicles, it becomes computationally challenging for platoons due to the exponentially growing action space. To address this, a safety-priority-based simulation model is proposed, which prioritizes vehicles with the smallest safety margins and conducts safety simulations in the twin world, ensuring efficient real-time computation and platoon cohesion.

Safety priority is determined based on headway time, with higher priority assigned to vehicles with smaller headway times relative to HDVs. The safety priority \( p_i \) is defined as:

\begin{equation}
    p_i = -\log \frac{d_{headway}}{v_t} + \sigma_i
\end{equation}
where \( d_{headway} \) is the headway distance, \( v_t \) is the current vehicle speed, and \( \sigma_i \) is a small random term to avoid identical safety priorities.

During the safety simulation, the vehicle with the highest safety priority undergoes assessment at each time step \( t \). The twin world checks whether the vehicle's decisions over the next \( T_n \) time steps conflict with nearby vehicles. If no collision is detected (i.e., the minimum distance between vehicles exceeds the threshold), the DRL decision is deemed safe. Otherwise, the twin world traverses other possible decisions, selecting the action that maximizes safety margins:

\begin{equation}
    a_t' = \arg\max_{a_t \in \mathcal{A}_{\mathrm{pro}}} \left( \min_{k \in T_n} d_{\mathrm{sm}, k} \right)
\end{equation} 
where \( \mathcal{A}_{\mathrm{pro}} \) represents alternative actions output by the DRL, and \( d_{\mathrm{sm}} \) refers to the safety margin, defined as the minimum headway distance in both current and target lanes during a left or right turn.
This risk-prioritized sequential prediction mechanism significantly reduces computational overhead and enables the safety prediction module to complete each decision in a shorter time.
The detailed procedure is illustrated in Algorithm \ref{alg:twin-world algorithm}.

\begin{algorithm}
\label{alg:twin-world algorithm}
    \SetAlgoLined
	\caption{Twin World Security Projection}
	\KwIn{$vehicles$,$x_{i}$,$y_{i}$,$v_x^{i}$,$v_y^{i}$,$headway_i$}
	\KwOut{$a_{i, i \in formation}$}

	\For{$i = 0$ to $N$}{
		Compute safety priority:
            \(priority_i = -\ln\left({headway_i}/{v_i}\right)\)\;
            Sort vehicles by safety priority and store in list $\phi$\;
	}
	\For{$j = 0$ to $|\phi|$}{
            Retrieve the vehicle with the highest safety priority: $\phi[0]$\;
            Identify neighboring vehicles and store them in list $N_{\phi[0]}$\;
            Predict the trajectory for the next $T_n$ simulation steps: $\xi_v, v \in \phi[0] \cup N_{\phi[0]}$\;
            \If{trajectories overlap}{
                Calculate safety margin $d_{\text{sm},i}$\;
                \If{lane changed}{
                    \(d_{\text{sm},i} = \min_j \left( p_i - p_j, j \in \text{objective} \, v_i \right)\) 
                }
                \Else{
                    \(d_{\text{sm},i} = \text{headway} \, v_i\)
                }
            Replace action $a_t \gets a_t'$\;
            Update trajectory list $\xi_{\phi[0]} \gets \xi_{\phi[0]}'$\;
            }
        Replace action $a_t \gets a_t'$
        Update $\phi[i] \gets \phi[i+1], i = 1, 2, \cdots$
        }
	\textbf{return} $a_{i, i \in formation}$
\end{algorithm}

\subsection{Formation maintenance}
\label{Formation maintenance}

\begin{algorithm}[htp]
\label{FSM}
    \SetAlgoLined
	\caption{Adaptive Switching Mechanism}
	\KwIn{$vehicles$, $riskLevel$, $Participants$, $L_{safe}$}
        \KwOut{$DecisionStrategy$}

	\For{$i = 0$ to $N$}{
        Assess risk level and number of traffic participants within $L_{safe}$\;
        \eIf{($riskLevel >= Routine Safe$) \textbf{and} (Participants within $L_{safe}$)}{
            Switch to State $S_1$: LQR Control Model\;
            Activate LQR Control\;
            Set $DecisionStrategy = LQR$\;
        }{
            Switch to State $S_2$: Safety-Enhanced Data-Driven Strategy\;
            Set $DecisionStrategy = DataDriven$\;
        }
        Execute $DecisionStrategy$ based on the current state\;
        \If{$DecisionStrategy == LQR$}{
            \textit{DecideWithLQR()}\;
        }
        \ElseIf{$DecisionStrategy == DataDriven$}{
            \textit{DecideWithDataDrivenStrategy()}\;
        }
    }
    \textbf{return} $DecisionStrategy$
\end{algorithm}

The platoon reconfiguration model ensures safe, robust decision-making in complex, high-interference scenarios but struggles with maintaining close headway distances in routine conditions due to conservative safety strategies. To address this, an adaptive switching mechanism, the platoon maintenance model, is proposed, incorporating optimal control methods.

This model maintains close headway in routine scenarios while dynamically adjusting strategies based on real-time traffic conditions and interference levels, enhancing both safety in complex scenarios and efficiency in routine conditions.

To allocate decision strategies across scenarios, a Finite State Machine (FSM) \( M(Q, D, \delta, q_0, F) \) is used, where:

\begin{enumerate}
    \item \( Q = \{S_1, S_2\} \) is the set of states;
    \item \( q \in Q \) represents the current state;
    \item \( \delta \) is the state transition function;
    \item \( F \) is the set of output states
    \item \( D \) denotes the risk level of the traffic scenario
\end{enumerate}

This study considers a scenario safe when the platoon is on the same lane and there are no other traffic participants within $L_{safe}$ ahead or behind the platoon. In this case, the FSM switches to state \( S_1 \), adopting the LQR control model\cite{9204679} to ensure smaller headway distances.
In all other scenarios, the FSM switches to state \( S_2 \), which employs a safety-enhanced data-driven strategy to balance the safety and efficiency of the platoon.
By defining these states and transitions, the FSM provides a structured approach to adaptively manage the decision strategies of the platoon, ensuring optimal performance across varying traffic conditions. The detailed procedure is illustrated in Algorithm \ref{FSM}.

\section{Simulation and Performance Evaluation}
\label{Simulation and Performance Evaluation}
In this section, the software-in-the-loop (SIL) experiments are conducted to validate the effectiveness of the proposed method. Additionally, a hardware-in-the-loop (HIL) experimental platform is built to further verify the model's safety and robustness and to assess its feasibility for direct deployment in real vehicles.

\subsection{Software-in-the-loop experiment}

\subsubsection{Environment and simulation setting}

\begin{figure*}[htp]
    \centering
    \includegraphics[width=\linewidth]{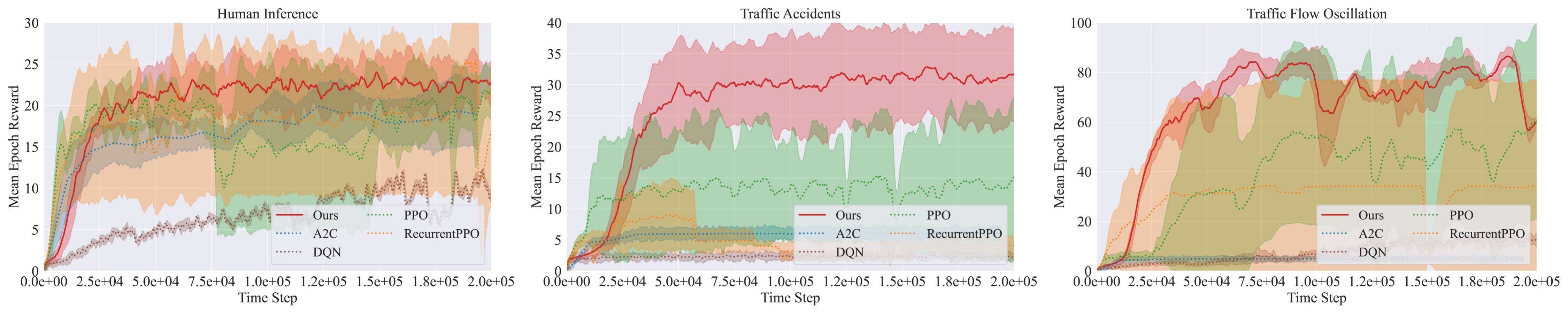}
    \caption{Schematic representation of the model training reward function in multiple scenarios.}
    \label{reward}
\end{figure*}

The simulation environment is constructed upon the Highway-Env simulator\cite{highway-env}, a reinforcement learning experimental environment library based on an OpenAI Gym environment. As illustrated in Figure \ref{schematics of the system}, all vehicles' high-level policies are transformed into throttle and steering information through a low-level PID controller, thereby enabling vehicle movement within the simulation environment. Furthermore, our HDVs employ the IDM\cite{treiber2000congested} car-following model and the MOBIL\cite{kesting2007general} lane-changing model provided by Highway-Env to achieve effective control in both longitudinal and lateral directions.

During the training process, a varying number of HDVs will randomly appear at randomly generated refresh points. The sampling frequency is set to \(15\) Hz, meaning decisions and controls are made 15 times per second. Meanwhile, to comprehensively evaluate the framework's limits against unscripted interactive adversarial conflicts, , supplementary stress tests based on Level-K game theory have also been conducted with training still remains on IMD+MOBIL, whoes detailed analytical setups and quantitative results available on \href{https://perfectxu88.github.io/Towards/}{our project website}.

Three different random seeds are used to train the proposed model in various complex scenarios, with each session running for over 100,000 steps until convergence. Four algorithms—DQN \cite{mnih2013playing}, A2C \cite{mnih2016asynchronous}, PPO \cite{schulman2017proximal}, and RecurrentPPO \cite{pleines2022generalization}—are selected as experimental baselines. The training parameters for the DRL model are detailed in Table \ref{tab:hyperparameter}.

\begin{table}[htp]
    \centering
    \caption{Hyperparameter settings for reinforcement learning}
    \begin{tabular}{ccc}
    
    \hline
    Param & Definition & Value \\
    \hline
        \(\eta\)& Learning Rate & \(5\times{10}^{-4}\)\\
        \(N_{steps}\)& steps for each environment per update & \(256\) \\
        \(N_t\) & Total Training Steps & \(10^{5}\) \\
        \(\beta_1\) & Loss function weights & 1 \\
        \(\beta_2\) & Loss function weights & 0.01 \\
        \(\lambda\) & Minibatch size & 128 \\
        \(\gamma\) & Discount factor & 0.8 \\  
    \hline
    \end{tabular}
    \label{tab:hyperparameter}
\end{table}

\subsubsection{Performance evaluation}

This section evaluates the performance of the proposed model in a simulated environment designed to reflect high-risk and safety-critical scenarios. As illustrated in Figure~\ref{fig:traffic scenario}, we construct three different scenarios, abstracted from real-world observations and intended to evaluate the robustness, adaptability, and safety of the decision-making framework.

Human interference simulates dynamic interactions with unpredictable HDVs, which are categorized as aggressive, neutral, and conservative, with behavior profiles sampled from distinct parameter distributions. Sudden braking events are introduced through rule-based deceleration patterns to represent common human-induced disturbances, such as cut-ins or emergency stops.

Traffic accidents introduce static obstructions or stalled vehicles in random positions ahead of the platoon, requiring coordinated reconfiguration, emergency braking, and controlled dissolution. Multiple accident configurations are included across lanes and locations to evaluate flexibility in reorganization strategies.

Traffic flow oscillations reflect congestion-driven instability caused by heterogeneous initial speeds and random vehicle ordering. A bimodal speed distribution is used to initialize traffic flow, generating stop-and-go waves that challenge longitudinal stability and coordination.

For all scenarios, initial vehicle states, including positions, speeds, and lane assignments, are sampled from uniform or bimodal distributions. HDVs are randomly placed across three lanes, and their longitudinal spacing is adapted to prevent collisions while promoting moderate traffic density. Each simulation run uses a unique random seed to ensure diverse traffic conditions during training and testing.
Each scenario class includes multiple parameterized variants to enable broad behavioral coverage and rigorous evaluation across both nominal and edge-case conditions\cite{sidorenko2021safety}.

\textbf{Comparison with DRL baselines.} Figure \ref{reward} illustrates the training performance of our model compared to baseline DRL algorithms across various scenarios. The training curves demonstrate that our model consistently outperforms the baseline algorithms, showing high adaptability and robustness essential for autonomous driving. This stability is attributed to our safety deduction model and innovative network structure, which enable our model to maintain reliable learning performance under challenging conditions and converge to a higher reward level.

Table \ref{rlcomparison} provides specific performance data. While our model may occasionally exhibit slightly lower speed and greater average headway distance compared to some baselines, it achieves significantly higher rewards, indicating that it prioritizes safety over raw efficiency. In contrast, the baseline algorithms reach their performance metrics at the expense of safety, which has been further verified in simulation experiments. Such trade-offs are unacceptable in transportation systems where safety is paramount.

In scenarios with high randomness and interference, such as traffic flow oscillations and traffic accidents, our model achieves significantly better convergence than other algorithms. This indicates an enhanced ability to learn optimal decisions under strong, dynamic disturbances, underscoring the robustness of our approach. 

\begin{table}[htp]
\centering
\caption{Performance metrics comparison between proposed method and benchmarks}
\begin{tabular}{ccccc}
\toprule
\textbf{Scenario} & \textbf{Model} & \textbf{Reward} & \textbf{Avg. Speed} & \textbf{Avg. HWD} \\
\midrule
\multirow{5}{*}{Sce-1} 
&A2C   & 18.93 & 22.23 & 10.00 \\
&DQN  & 9.47 & 25.18  & 20.06 \\
&PPO   & 22.10 & 23.83 & 13.85 \\
&RPPO  & 17.43 & 17.57  & 9.73 \\
&Ours   & \textbf{23.07} & 20.57 & 9.93 \\
\midrule
\multirow{5}{*}{Sce-2} 
&A2C   & 6.04 & 20.34 & 10.50 \\
&DQN  & 2.31 & 27.72  & 9.81 \\
&PPO   & 15.05 & 20.75 & 14.37 \\
&RPPO  & 4.10 & 25.70  & 17.10 \\
&Ours   & \textbf{31.66} & 20.59 & 11.33 \\
\midrule
\multirow{5}{*}{Sce-3} 
&A2C   & 4.87 & 20.18 & 10.00 \\
&DQN  & 11.83 & 20.71  & 12.18 \\
&PPO   & 58.21 & 20.13 & 10.00 \\
&RPPO  & 34.19 & 20.15  & 9.46 \\
&Ours   & \textbf{61.07} & 20.93 & 8.95 \\
\bottomrule
\end{tabular}
\label{rlcomparison}
\end{table}

\textbf{Comparison with Leading Platooning Algorithms.}
To evaluate the effectiveness of the TriCoD framework, a comparative analysis is conducted with several leading platooning algorithms. This analysis is based on four key performance indicators: average vehicle speed, average headway, collision rate, and pass rate. These metrics provide a comprehensive assessment of the model’s performance in terms of safety and efficiency. Notably, the pass rate metric is designed to consider scenarios where the convoy may need to halt to avoid hazards, thereby balancing risk avoidance with maintaining traffic flow.

In the experiments, the model is evaluated against advanced lane-changing algorithms, including Successive Platoon Lane Change (SuPLC)\cite{wang2022cut} and Simultaneous Platoon Lane Change (SiPLC)\cite{liu2023coordinated}, as well as a baseline DRL model across different scenarios: Human Inference, Traffic Accidents, and Traffic Flow Oscillation.
Specifically, SuPLC is a trajectory optimization-based control method that accounts for vehicle dynamics and guarantees string and lateral stability during staggered lane changes, while SiPLC employs a combinatorial optimization approach formulated as an Integer Linear Programming problem to generate coordinated lane change sequences across multi-lane scenarios.

To provide a more comprehensive safety assessment, we introduce Minimum Time to Collision (TTC) and Deceleration Rate to Avoid Crash (DRAC) as measures. TTC measures the time to collision between platoon vehicles and surrounding HDVs. A lower TTC indicates a higher collision risk, while a higher value suggests safer conditions: 

\begin{equation}
    TTC_{C_i,H_j} = \min_{j \in \text{HDVs}} \left( \frac{d_{C_i,H_j}}{v_{C_i} - v_{H_j}} \right)
\end{equation}
where $d_{C_i,H_j}$ is the distance between vehicle $C_i$ and HDV $H_j$, and $v_{C_i}$ and $v_{H_j}$ are their respective velocities. DRAC evaluates the minimum deceleration required to avoid a collision, which is calculated for each vehicle in the platoon by considering surrounding vehicles:

\begin{equation}
    DRAC_{C_i} = \max_{j \in \text{Vehicles}} \left( \frac{(v_{C_i} - v_j)^2}{2 \left( d_{C_i,j} - L \right)} \right)
\end{equation}
where $v_j$ represents the velocity of vehicle $j$, and $L$ is the length of the vehicle.

As shown in Table \ref{Performance metrics}, our model consistently outperformed these benchmarks in terms of collision rate and pass rate, demonstrating a unique balance between safety and operational efficiency.
While some baseline models achieved slightly higher speeds or tighter headway distances, they do so at the expense of safety, as indicated by increased collision rates. In contrast, the TriCoD model maintained a zero-collision rate and a perfect pass rate of 100\%, while still being able to maintain a relatively high pass rate and a smaller average headway.
This balance illustrates our model's capacity to prioritize safety without sacrificing operational effectiveness, especially in scenarios involving high-risk factors like traffic accidents and flow oscillations.
In terms of safety, TriCoD outperformed the baseline models across all scenarios. It consistently achieved higher minimum TTC and lower DRAC, indicating superior collision avoidance and a more conservative, safer approach. These results demonstrate TriCoD’s ability to maintain stability and safety, even in challenging, fluctuating traffic conditions.

\textbf{Cross-Validation.}
To evaluate the adaptability and generalization capabilities of the learned strategies, we test the models in task environments beyond their training conditions. Specifically, we introduce distribution shift by transferring policies across unseen traffic regimes with different densities and HDV ratios (Simple/Medium/Hard: 10/20/30 veh/km/lane). As shown in Table~\ref{Cross_validation}, our method preserves stable returns and consistently low collision rates across all cross-level evaluations, demonstrating strong adaptability to changes in interaction intensity and background traffic composition. In contrast, the DRL baseline exhibits marked performance degradation under distribution shift, with substantial reward collapse and sharply increased collision frequency as the applied traffic regime deviates from the training setting. These results support the generalization claim by evidencing that the proposed framework maintains both effectiveness and safety under unseen traffic distributions.

Overall, these results indicate that our model consistently achieves high safety and operational metrics across varied and challenging environments, demonstrating an effective balance that baseline models do not maintain.

\begin{table*}
\centering
\renewcommand{\arraystretch}{1.25}
\caption{Performance metrics comparison of multiple metrics\\ between the proposed method and three state-of-the-art benchmarks}
\begin{tabular}{ l l c c c c c c}
\toprule
\textbf{Scenario} & \textbf{Model} & \textbf{Avg. Speed} & \textbf{Avg. HWD} & \textbf{Min. TTC} & \textbf{Max. DRAC} & \textbf{Collision Rate} & \textbf{Pass Rate} \\
\hline
\multirow{4}{*}{Human Inference} & SuPLC & 22.58 & 9.68 & 0.00 & 6.00 & 1.00 & 0.00 \\
 & SiPLC & 22.58 & 9.68 & 0.00 & 6.00 & 1.00 & 0.00 \\
 & DRL & 20.46 & 10.00 & 5.80 & 0.25 & 0.00 & 1.00 \\
 \rowcolor{gray!10}
 & Ours & 20.57 & \textbf{9.93} & \textbf{7.15} & \textbf{0.13} & \textbf{0.00} & \textbf{1.00} \\
\hline
\multirow{4}{*}{Traffic Accident} & SuPLC & 21.96 & 10.18 & 2.75 & 0.51 & 0.52 & 0.29 \\
 & SiPLC & 20.81 & 10.15 & 4.69 & 0.23 & 0.54 & 0.21 \\
 & DRL & 21.13 & 14.10 & 1.50 & 0.90 & 0.19 & 0.91 \\
  \rowcolor{gray!10}
 & Ours & 20.59 & \textbf{11.33} & 1.58 & 0.82 & \textbf{0.00} & \textbf{1.00} \\
\hline
\multirow{4}{*}{Traffic Flow Oscillation} & SuPLC & 20.28 & 10.13 & 1.47 & 1.05 & 0.03 & 0.96 \\
 & SiPLC & 19.68 & 10.15 & 1.73 & 0.93 & 0.03 & 0.97 \\
 & DRL & 20.93 & 10.08 & 2.60 & 1.40 & 0.09 & 0.91 \\
  \rowcolor{gray!10}
 & Ours & \textbf{20.93} & \textbf{8.95} & \textbf{3.19} & \textbf{0.44} & \textbf{0.00} & \textbf{1.00} \\
\bottomrule
\end{tabular}
\label{Performance metrics}
\end{table*}

\begin{table}[t]
\centering
    \caption{Performance under different difficulty transfer settings.}
    \begin{tabular}{cccccc}
    \toprule
    \multirow{2}{*}{\makecell{\textbf{Original}\\\textbf{Level}}} &
    \multirow{2}{*}{\makecell{\textbf{Applied}\\\textbf{Level}}} &
    \multicolumn{2}{c}{\textbf{Evaluation Reward}} &
    \multicolumn{2}{c}{\textbf{Collision Rate}} \\
    
    \cmidrule(lr){3-4}\cmidrule(lr){5-6}
    & & Ours & DRL & Ours & DRL \\
    \midrule
    
    \multirow{3}{*}{Hard}  & Simple & 54.73 & 50.24 & 0.00 & 0.12 \\
                           & Medium & 53.49 & 47.55 & 0.00  & 0.17  \\
                           & Hard & 54.70 & 16.87 & 0.02 & 0.78 \\
    \midrule
    
    \multirow{3}{*}{Medium} & Simple & 54.60 & 46.75 & 0.00 & 0.15  \\
                           & Medium & 51.64 & 50.72 & 0.00 & 0.09 \\
                           & Hard & 49.57 & 4.17 & 0.05 & 1.00  \\
    \midrule
    
    \multirow{3}{*}{Simple} & Simple & 53.56 & 47.73 & 0.03 & 0.18 \\
                           & Medium & 49.83 & 47.05 & 0.07 & 0.27 \\
                           & Hard & 35.48 & 6.47 & 0.37 & 1.00 \\
    \bottomrule
    \end{tabular}
    \label{Cross_validation}
\end{table}

\subsection{Hardware-in-loop experiment}

\begin{figure}
    \centering
    \subfigure[HIL experimental platform framework]{
    \label{HIL experimental platform framework.}
    \includegraphics[width=0.9\linewidth]{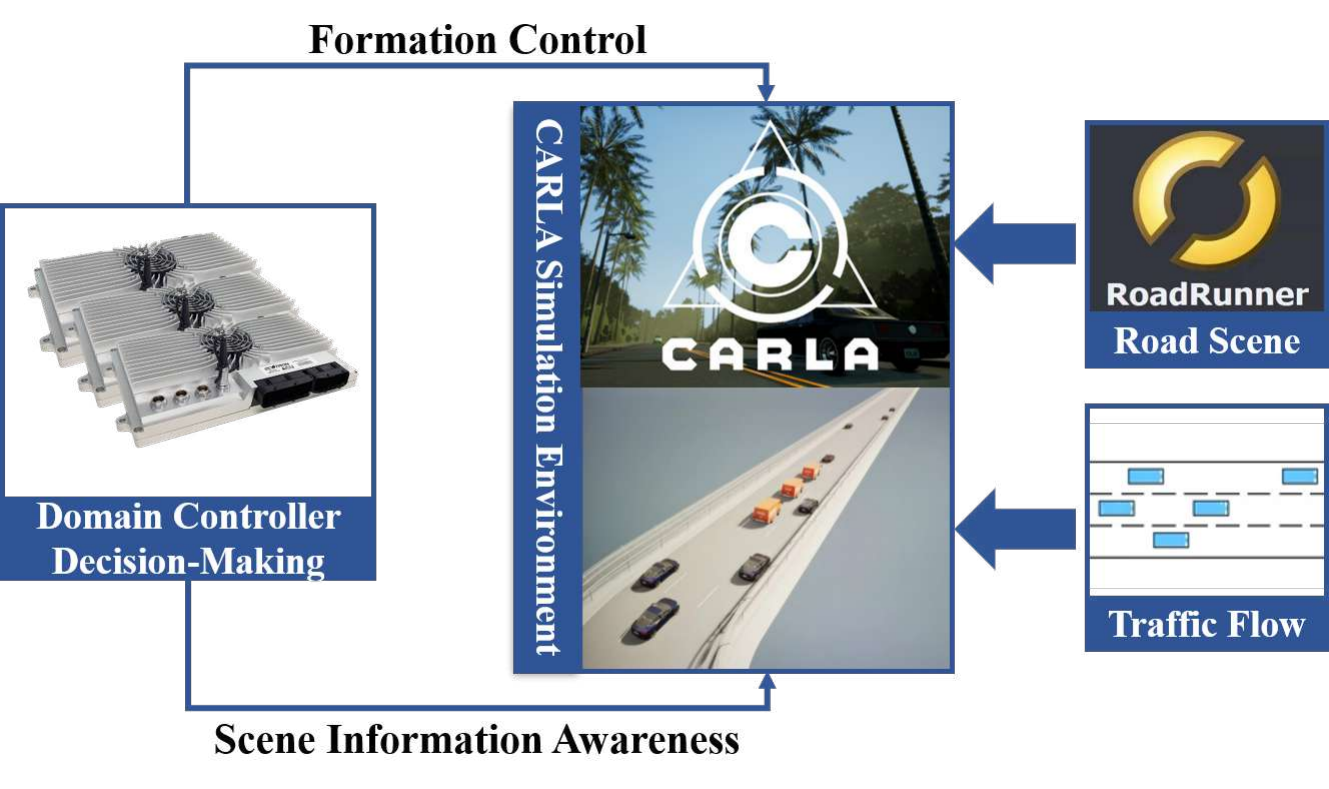}
    }
    \subfigure[HIL experiment platform physical diagram]{
    \label{HIL experiment platform physical diagram}
    \includegraphics[width=0.8\linewidth]{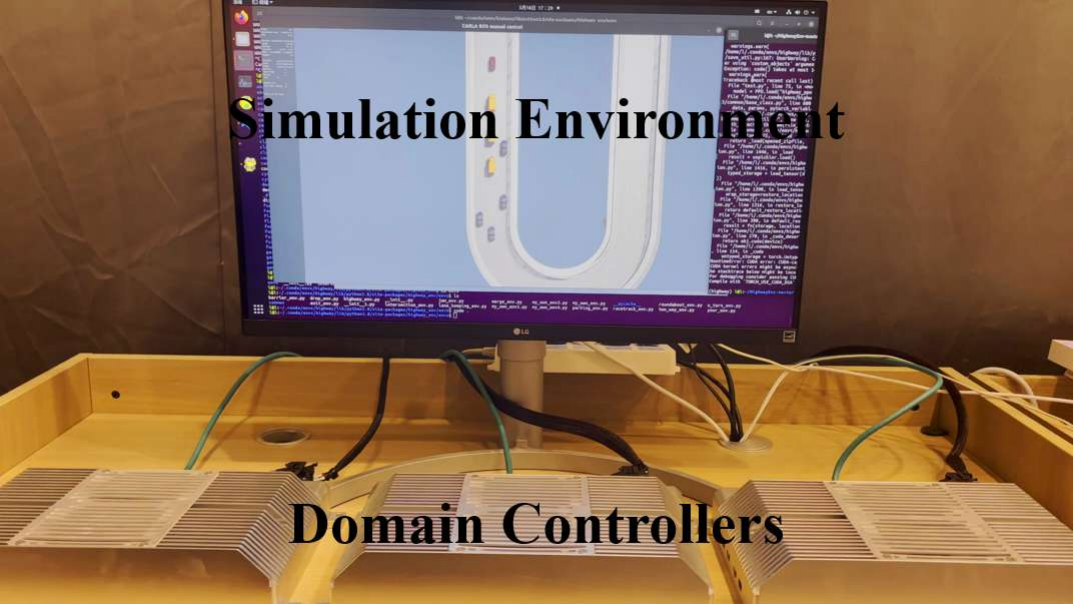}
    }
    \caption{Illustration of the HIL platform architecture for autonomous vehicle platooning, which integrates a CARLA simulation environment, RoadRunner for road scenes and traffic flow, and domin controllers for decision-making and control.}
    \label{Overall Architecture of the HIL Experimental Platform}
\end{figure}

\begin{figure*}[htp]
    \centering
    \includegraphics[width=\linewidth]{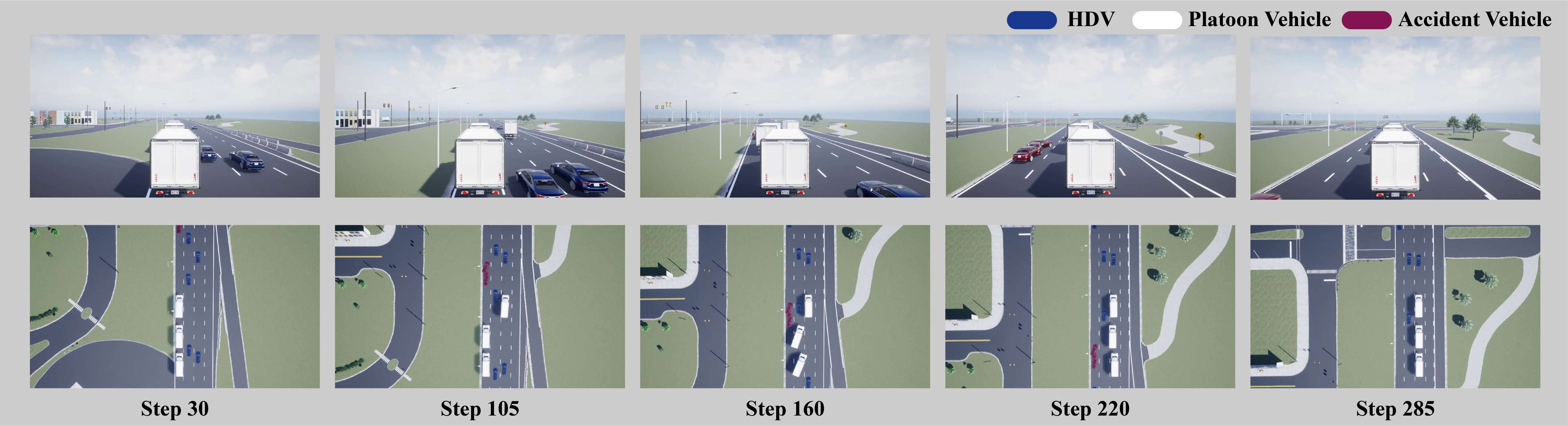}
    \caption{Illustration of the adaptive response to static obstruction in autonomous vehicle platooning, which decipts a scenario where the platoon encounters a multi-vehicle collision, necessitating coordinated lane changes to navigate the obstruction safely.}
    \label{Barrier_case}
\end{figure*}

\begin{figure*}[htp]
    \centering
    \includegraphics[width=\linewidth]{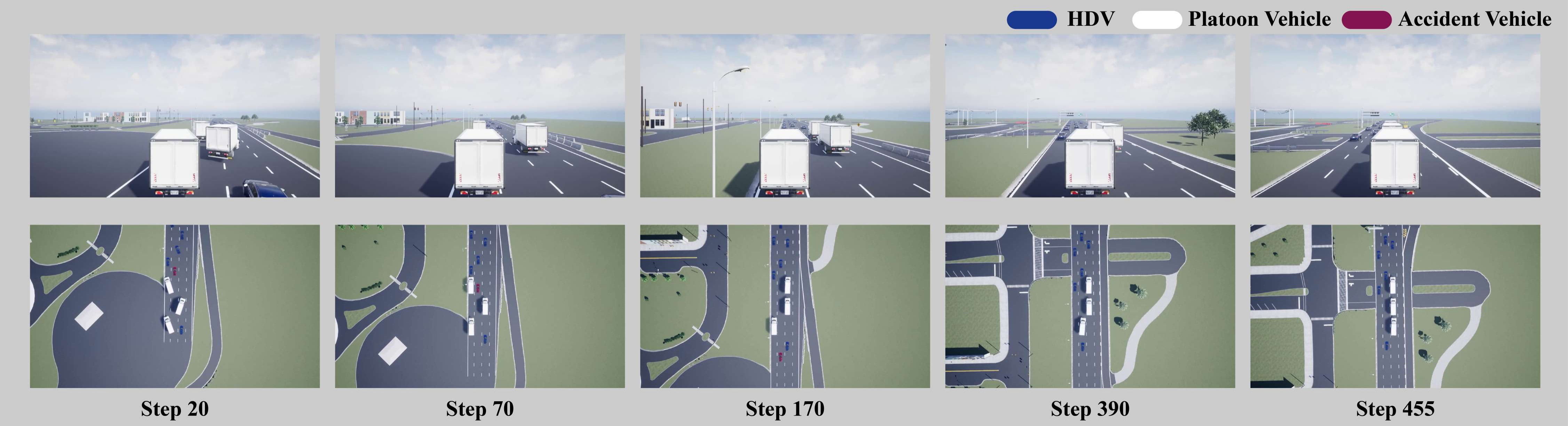}
    \caption{Illustration of the adaptive response to dynamic obstruction in autonomous vehicle platooning, which depicts a scenario where the platoon leader's sudden deceleration triggers a staggered lane-change maneuver among following vehicles to prevent collisions and maintain traffic flow.}
    \label{Drop_case}
\end{figure*}

\subsubsection{HIL experimental platform construction}
In the HIL experiments, an advanced simulation platform is constructed by integrating a domain controller with Carla \cite{Dosovitskiy17}, a high-fidelity driving simulator, to validate the proposed model in a controlled yet realistic environment. The experimental highway scenario is created using Roadrunner, with varied background traffic flows generated by Highway-env.

The platform consists of the domain controller, Carla simulation environment, Roadrunner, Highway-env, and PID controllers. The domain controller, whose parameters are shown in Table \ref{tab:test_performance}, processes real-time traffic and road information from Carla. Carla provides detailed road conditions and vehicle dynamics, which are translated into vehicle movements via PID controllers based on high-level action commands. Roadrunner generates complex highway scenarios, while Highway-env produces traffic with varied driving styles to test the algorithm's robustness.

During experiments, the domain controller gathers real-time data from Carla, including vehicle positions, velocities, and traffic conditions.Based on this data, the proposed DRL algorithm makes high-level decisions, such as speed adjustments and lane changes. These decisions are sent back to Carla, where PID controllers convert them into specific throttle and steering commands. Various scenarios, including high-risk and normal conditions, are simulated to evaluate the algorithm's performance based on metrics like safety, robustness, and efficiency. Figure \ref{Overall Architecture of the HIL Experimental Platform} illustrates the platform's overall architecture, highlighting its key components.

\begin{table}
    \centering
    \caption{Specifications of the domain controller}
    \begin{tabular}{cc}
         \hline
         Parameter& Specification \\
         \hline
         CPU& Cortex-A78AE – Arm®\\
         CPU Frequency&2.2 GHz \\
         Memory& 64 GB 256-bit 204.8 GB/s\\
         GPU Performance&275 TOPS \\
         Storage& 2 TB SSD\\
         \hline
    \end{tabular}
    \label{tab:test_performance}
\end{table}

\subsubsection{Case analysis}
This section examines the performance of our model on the HIL simulation platform in real-time autonomous platooning scenarios. The results indicate that our model can effectively perform perception and decision-making, ensuring safe and robust platooning.
It is worth noting that in the experiments on the HIL platform, the average decision-making time of our model is 6ms, which verifies that the model is suitable for practical deployment in rapidly changing traffic environments.

Two typical scenarios with significant static and dynamic interferences are analyzed to illustrate the model's strategies. Further demonstration videos are available on our website \footnote{\href{https://perfectxu88.github.io/Towards/}{Hardware-in-the-Loop Experimental Validation Video Weblink}}.

In the first scenario, the platoon encounters a multi-vehicle collision blocking several lanes. Traditional strategies struggle with safe deceleration and coordination, but our model enables the lead vehicle to initiate a lane change, followed by others when safe. As shown in Figure \ref{Barrier_case}, the lead vehicle moves at step 30, with all vehicles clearing the obstruction by step 220. The LQR controller stabilizes the platoon with reduced headway, illustrating flexibility in static obstructions. In the second scenario, the lead vehicle suddenly decelerates at 5 m/s² due to a breakdown. Traditional models risk collision due to short distances, but ours initiates a staggered lane change (see Figure \ref{Drop_case}). By step 70, all vehicles are safely repositioned, and the platoon reforms in the right lane by step 455, maintaining safety and flow.

These scenarios demonstrate the model's capability to handle sudden obstacles by dynamically adjusting the platoon's formation and leveraging available lane space to maintain safety and efficiency.

\section{Conclusion}
\label{Conclusion}
In the evolving field of autonomous driving, ensuring safe and robust self-organizing decision-making for vehicle platooning, especially in mixed traffic, presents significant challenges. This paper introduces TriCoD, a Twin-World Safety-Enhanced Data-Model-Knowledge Triple-Driven Adaptive Autonomous Vehicle Platooning Cooperative Decision-Making Framework. By integrating reinforcement learning with model control techniques, TriCoD enables adaptive decision-making across diverse, complex scenarios, guided by traffic priors and optimal control strategies.

The proposed model significantly outperforms traditional platooning methods, particularly in safety, efficiency, and robustness.
Extensive SIL and HIL experiments validate the effectiveness of the proposed framework in handling structured static and dynamic disturbances.
Future work will focus on developing closed-loop, data-driven adversarial scenario generation to systematically evaluate emergent interactive conflicts, as well as optimizing the model for scalability in larger platoons and integrating additional sensory data to enhance situational awareness.
These advancements will further advance the reliability and applicability of autonomous platooning systems in commercial deployments.

\ifCLASSOPTIONcaptionsoff
  \newpage
\fi

\bibliographystyle{unsrt}
\bibliography{bibtex/bib/related}

\begin{IEEEbiography}[{\includegraphics[width=1in,height=1.25in,clip,keepaspectratio]{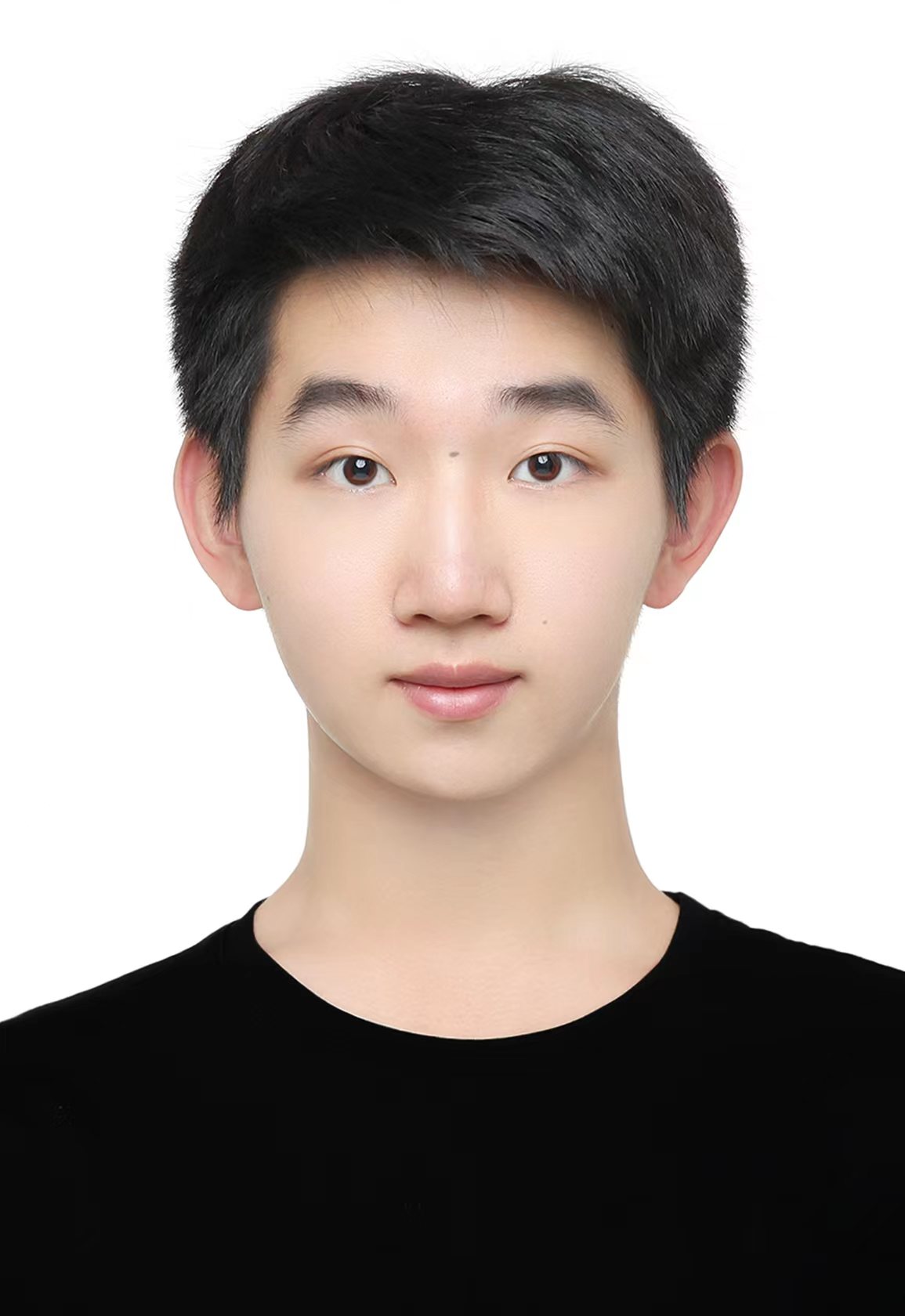}}]{Chengkai Xu} (Member, IEEE) received the B.S. degree in Transportation Engineering from Tongji University, where he is currently pursuing the M.S. degree. His research interests include end-to-end autonomous driving, generative artificial intelligence, reinforcement learning, and embodied intelligence.
\end{IEEEbiography}\vspace{-6pt}

\begin{IEEEbiography}[{\includegraphics[width=1in,height=1.25in,clip,keepaspectratio]{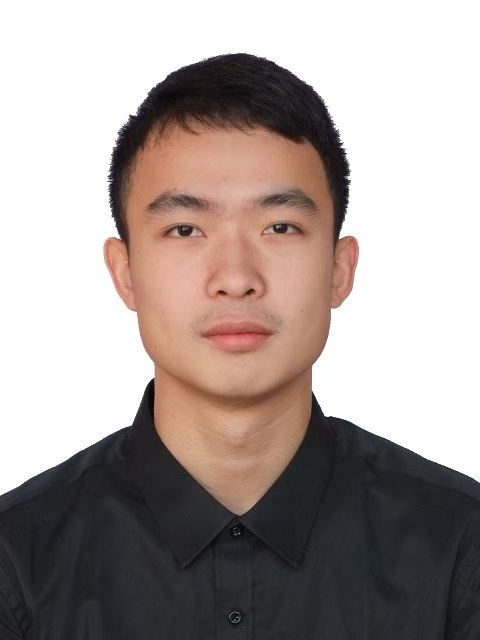}}]{Zihao Deng} received the B.S. degree in Microelectronics Science and Engineering from Tongji University, China, in 2025. He is currently pursuing the M.S. degree in Pattern Recognition and Intelligent Systems with the State Key Laboratory of Multimodal Artificial Intelligence Systems, Institute of Automation, Chinese Academy of Sciences, China. His research interests include multimodal learning, auto-formalization, and autonomous driving.
\end{IEEEbiography}\vspace{-6pt}

\begin{IEEEbiography}[{\includegraphics[width=1in,height=1.25in,clip,keepaspectratio]{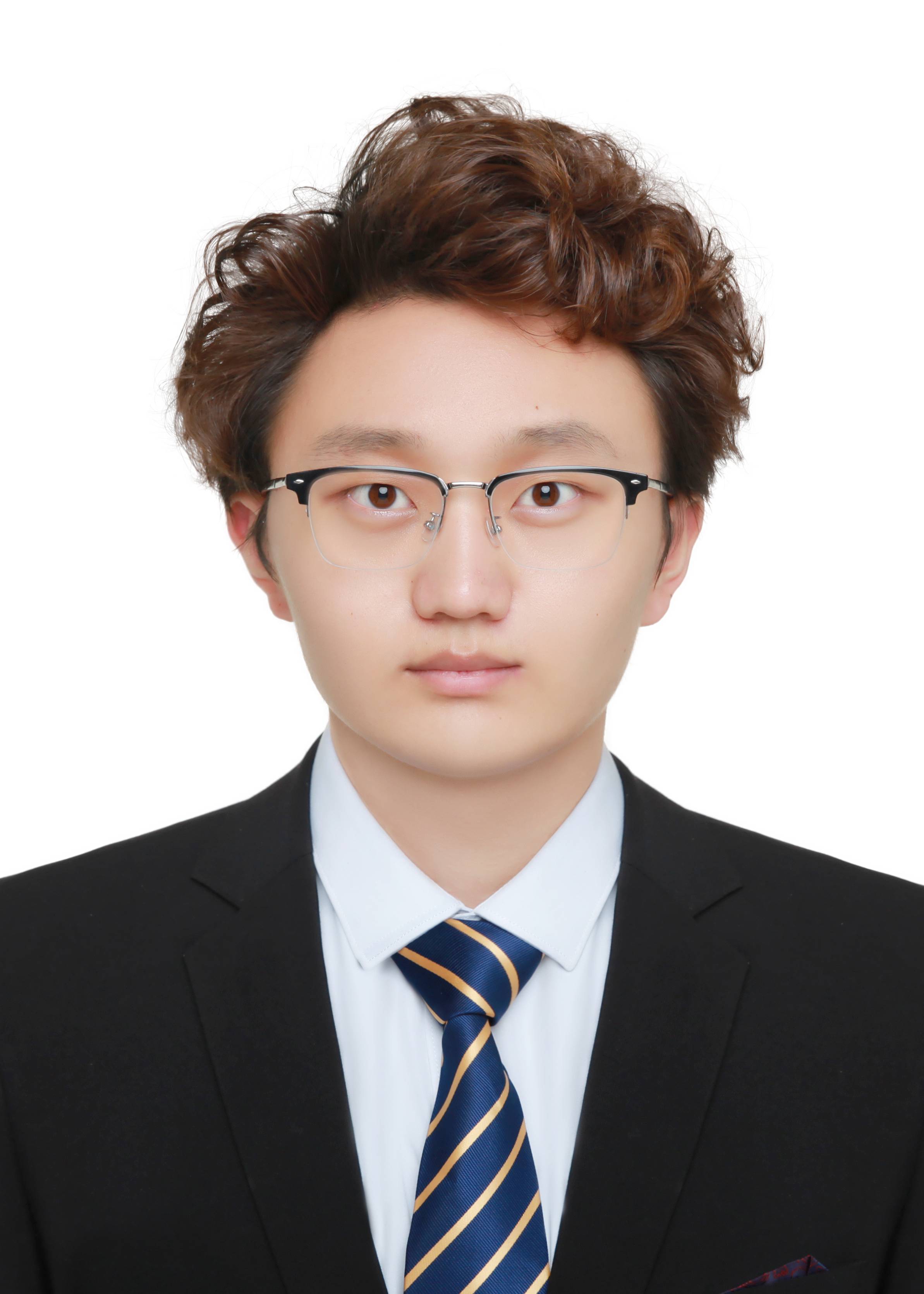}}]{Jiaqi Liu} (Graduate Student Member, IEEE) is a PhD  student  in the Department of Computer Science at UNC-Chapel Hill. He received the B.S. and M.S. degree from Tongji University. His research interests include Embodied AI, multimodal LLMs, autonomous driving and reinforcement learning.  He was a Visiting Researcher with the Department of Mechanical Engineering, University of California, Berkeley.
\end{IEEEbiography}\vspace{-6pt}

\begin{IEEEbiography}[{\includegraphics[width=1in,height=1.25in,clip,keepaspectratio]{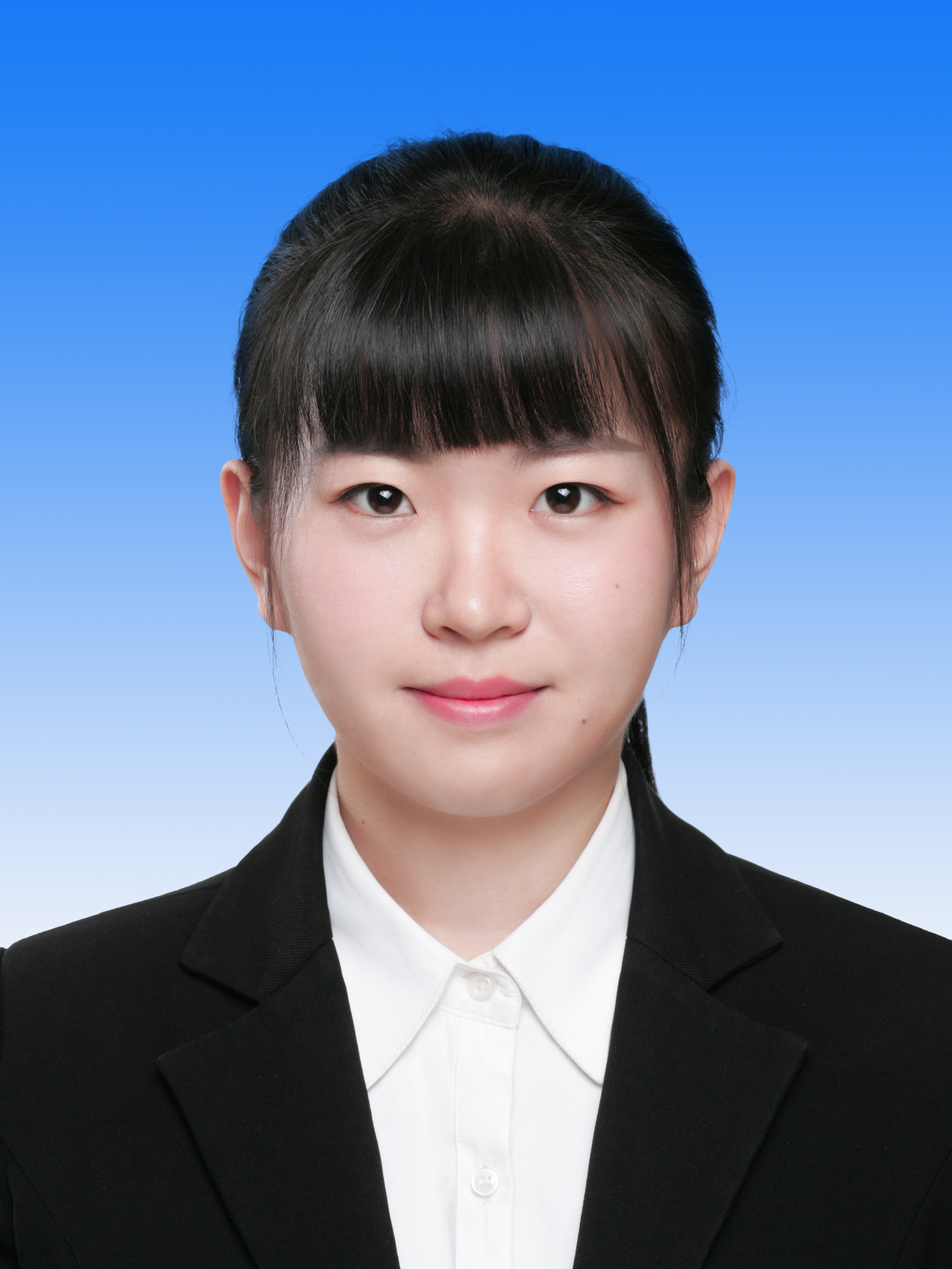}}]{Aijing Kong} received the B.S. degree in Automotive Engineering from Harbin Institute of Technology (Weihai) in 2020. She is currently pursuing the Ph.D. degree at Tongji University. Her research interests include decision-making and control for autonomous driving, cooperative platoon control for autonomous vehicles, and multi-agent systems.
\end{IEEEbiography}\vspace{-6pt}

\begin{IEEEbiography}[{\includegraphics[width=1in,height=1.25in,clip,keepaspectratio]{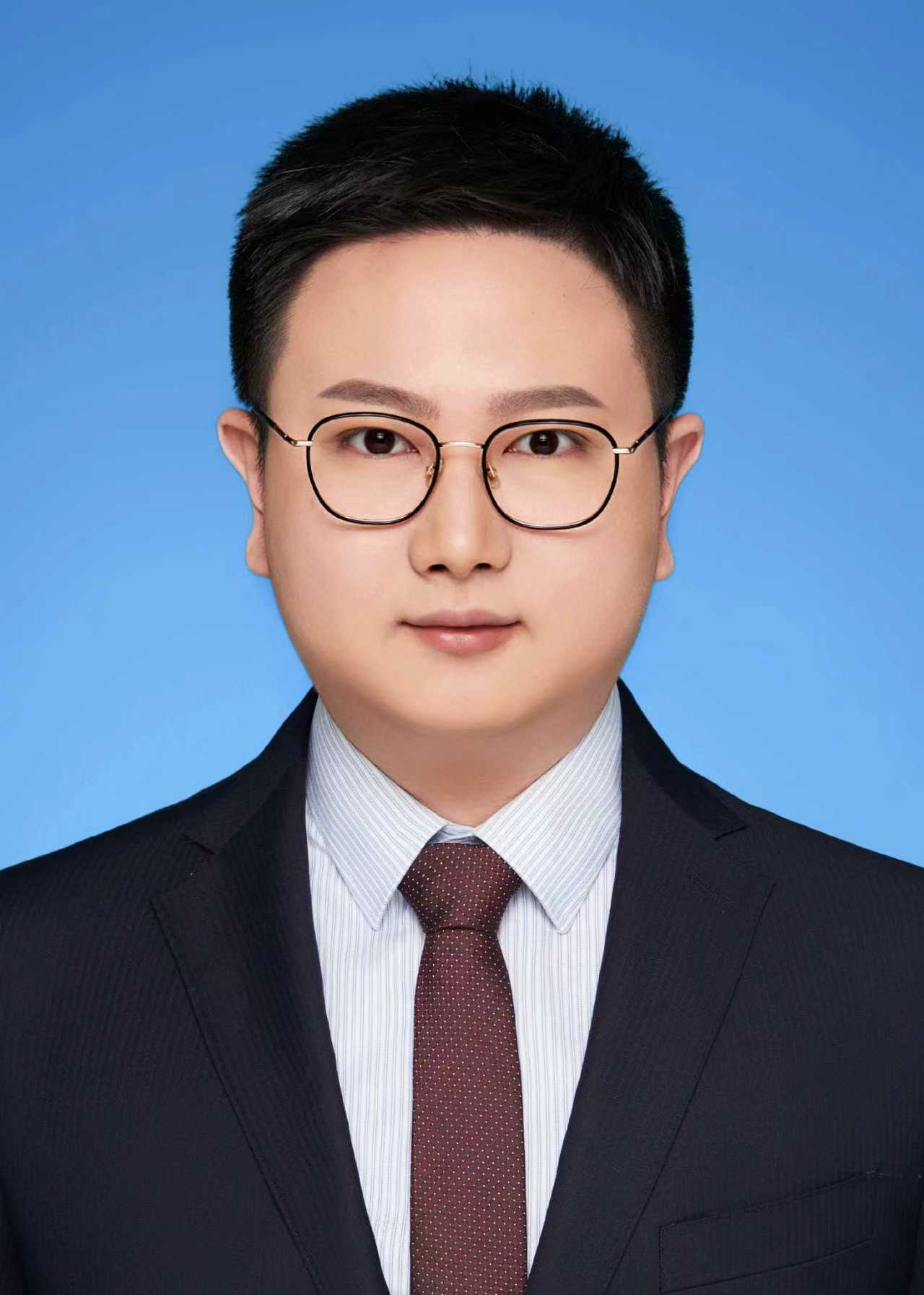}}]{Yu Tang} is a Director of the Intelligent Cockpit Evaluation Department at China Automotive Engineering Research Institute Co., Ltd., specializing in ICV testing and evaluation technologies with a focus on multi-pillar testing of intelligent driving in complex environments, typical scenario construction, and anthropomorphic evaluation. He received the B.S. degree from Beijing Institute of Technology in 2014 and M.S. degree at the Karlsruhe Institute of Technology.
He has won the first prize of China-SAE Science and Technology Award, the first prize of ITS-China Science and Technology Award, the first prize of the BRICS Industrial Innovation Contest, the Gold Medal at the International Exhibition of Inventions Geneva.

\end{IEEEbiography}\vspace{-6pt}

\begin{IEEEbiography}[{\includegraphics[width=1in,height=1.25in,clip,keepaspectratio]{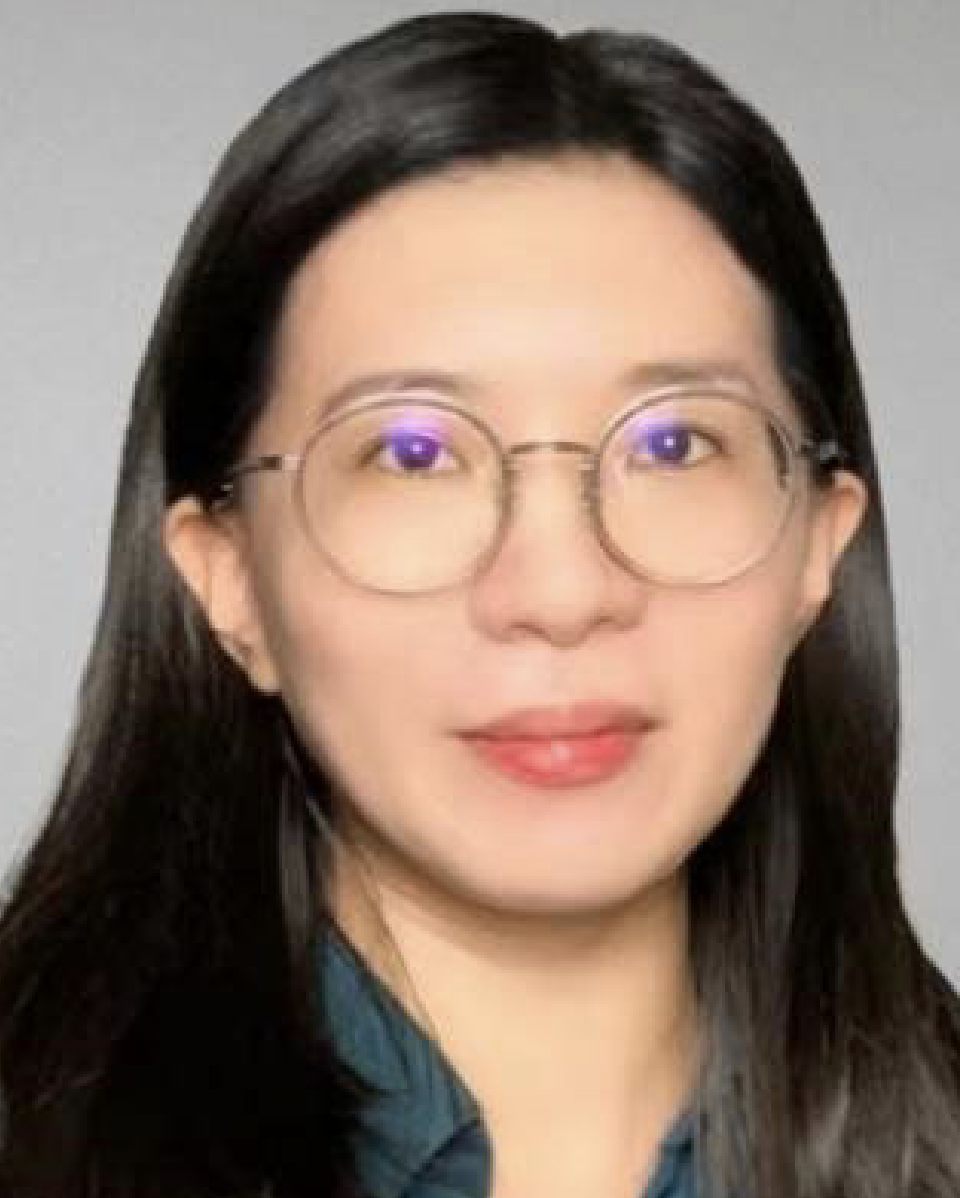}}]{Chao Huang} (Senior Member, IEEE) received the Ph.D. degree from the University of Wollongong, Wollongong, NSW, Australia, in 2018. She is currently a Research Assistant Professor with the Department of Industrial and Systems Engineering, The Hong Kong Polytechnic University, Hong Kong. Her research interests are humanmachine collaboration, mobile robots, and path planning. Dr. Huang serves as an Associate Editor for IEEE TRANSACTIONS ON TRANSPORTATION ELECTRIFICATION, Engineering Applications of Artificial Intelligence, and Control Engineering Practice.
\end{IEEEbiography}\vspace{-6pt}

\begin{IEEEbiography}[{\includegraphics[width=1in,height=1.25in,clip,keepaspectratio]{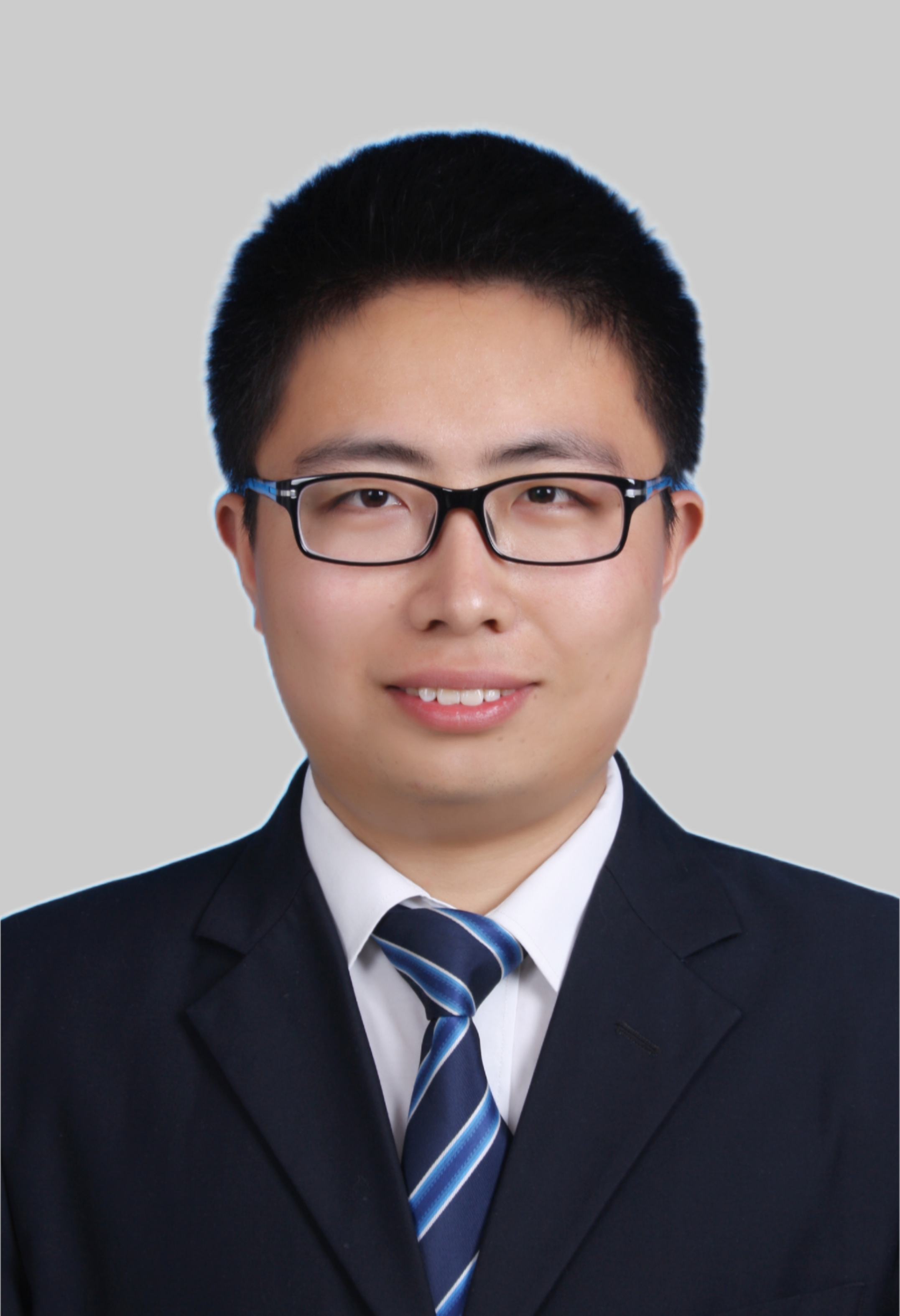}}]{Peng Hang} (Senior Member, IEEE) is a Research Professor at the Department of Traffic Engineering, Tongji University, Shanghai, China. He received the Ph.D. degree with the School of Automotive Studies, Tongji University, Shanghai, China, in 2019. He was a Visiting Researcher with the Department of Electrical and Computer Engineering, National University of Singapore, Singapore, in 2018. From 2020 to 2022, he served as a Research Fellow with the School of Mechanical and Aerospace Engineering, Nanyang Technological University, Singapore. His research interests include vehicle dynamics and control, decision making, motion planning and motion control for autonomous vehicles. He serves as an Associate Editor of IEEE Transactions on Vehicular Technology, Journal of Field Robotics, IET Smart Cities, and SAE International Journal of Vehicle Dynamics, Stability, and NVH.
\end{IEEEbiography}\vspace{-6pt}

\end{document}